\documentclass[review]{elsarticle}

\usepackage{lineno,hyperref}
\modulolinenumbers[5]

\journal{Artificial Intelligence}

\usepackage[left=2cm,right=2cm,top=2.0cm,bottom=2.0cm]{geometry}
\usepackage{fancyvrb}

\usepackage{verbatim}

\usepackage[fit]{truncate}
\usepackage{float}
\usepackage{amsthm}
\usepackage{appendix}
\usepackage{amsmath}
\usepackage{amssymb}
\usepackage{multicol}
\usepackage{multirow}
\usepackage[ruled]{algorithm2e}
\usepackage{amsfonts}
\usepackage{algpseudocode}
\usepackage{array}
\usepackage[numbers]{natbib}
\usepackage{enumitem}
\usepackage{changepage}
\usepackage{lipsum}
\usepackage{framed}
\usepackage{xcolor}
\colorlet{shadecolor}{gray!40}
\usepackage{setspace}
\usepackage{xspace}
\usepackage{titlecaps}
\usepackage{bibentry}
\nobibliography*
\usepackage[most]{tcolorbox}

\usepackage{pdfpages}
\usepackage{txfonts}
\usepackage{pxfonts}
\usepackage{listings}
\usepackage{pgfplots}
\usepackage{subfig}
\usepackage{tabularx}
\usepackage{bbm}

\Addlcwords{as is for an of and to at if the from with a each in on into -- *}

\newenvironment{acknowledgements}
{
  \cleardoublepage

  \thispagestyle{plain}

  \begin{center}
    {\large \bf Acknowledgements}
  \end{center}

}{\cleardoublepage}

\newenvironment{dedication}
{
  \cleardoublepage

  \thispagestyle{plain}

  \begin{center}
    {\large \bf Dedication}
  \end{center}

  \begin{center}

}{\cleardoublepage}
\def\enddedication{
  \par
\end{center}

\cleardoublepage

}

\usepackage{lineno}

\usepackage[nopostdot, style=super, toc]{glossaries}

\usepackage{tikz}
\usetikzlibrary{shapes,chains,arrows.meta,calc,decorations.markings,math,arrows.meta,matrix, positioning, patterns}

\newtheorem{definitionx}{Definition}
\newenvironment{definition}
  {\definitionx}
  {\enddefinitionx}

\theoremstyle{definition}
\newtheorem{examplex}{Example}
\newenvironment{example}
  {\pushQED{\qed}\examplex}
  {\popQED\endexamplex}

\newcommand{\define}[1]{{\bf #1}}

\usepackage{thrmappendix}
\newtype{lemma}{theorem}{Lemma}

\makeatletter
\def\thmhead@plain#1#2#3{%
  \thmname{#1}\thmnumber{\@ifnotempty{#1}{ }\@upn{#2}}%
\thmnote{ {\the\thm@notefont#3}}}
\let\thmhead\thmhead@plain
\makeatother

\definecolor{shadecolor}{rgb}{0.95,0.95,0.95}

\makeatletter
\renewenvironment{abstract}{%
    \begin{center}%
      {\bfseries \abstractname\vspace{-.5em}\vspace{\z@}}%
    \end{center}%
  }
{}
\makeatother

\theoremstyle{definition}

\newcommand{\sequence}[1]{\textbf{#1, ...}}

\newcommand{\fork}{%
  \mathbin{%
    \supset
    \mathrel{\mkern-9mu}%
    \mathrel{-}%
  }%
}

\def\set#1{\{#1\}}

\newcommand{\bigsetbegin}{ \left\{ \begin{array}{l}}
\newcommand{\bigsetend}{ \end{array}\right\}}

\newcommand{\sys}{{\textsc{Apperception Engine}}}

\newcommand{\logic}{Datalog\textsuperscript{$\fork$}}

\newcommand{\defbegin}{
	 \begin{definition}
}
\newcommand{\defend}{
	 \end{definition}
}

\newtheorem{innercustomgeneric}{\customgenericname}
\providecommand{\customgenericname}{}
\newcommand{\newcustomtheorem}[2]{%
  \newenvironment{#1}[1]
  {%
   \renewcommand\customgenericname{#2}%
   \renewcommand\theinnercustomgeneric{##1}%
   \innercustomgeneric
  }
  {\endinnercustomgeneric}
}

\newcustomtheorem{customTheorem}{Theorem}

\let\OLDthebibliography\thebibliography
\renewcommand\thebibliography[1]{
  \OLDthebibliography{#1}
  \setlength{\parskip}{0pt}
  \setlength{\itemsep}{0pt plus 0.3ex}
}

\begin{document}

\begin{frontmatter}

\title{Evaluating the Apperception Engine}

\author[address-2-deepmind,address-1-imperial]{Richard Evans\corref{mycorrespondingauthor}}
\cortext[mycorrespondingauthor]{Corresponding author}
\ead{richardevans@google.com}

\author[address-3-valencia,address-4-cambridge]{Jos\'e Hern\'andez-Orallo}
\author[address-2-deepmind]{Johannes Welbl}
\author[address-2-deepmind]{\mbox{Pushmeet Kohli}}
\author[address-1-imperial]{Marek Sergot}

\address[address-2-deepmind]{DeepMind, London}
\address[address-1-imperial]{Imperial College London}
\address[address-3-valencia]{Universitat Polit\`ecnica de Val\`encia}
\address[address-4-cambridge]{CFI, University of Cambridge}

\end{frontmatter}



\begin{abstract}

The \sys{} \cite{evans2019making} is an unsupervised learning system. 
Given a sequence of sensory inputs, it constructs a symbolic causal theory that both explains the sensory sequence and also satisfies a set of unity conditions.
The unity conditions insist that the constituents of the theory -- objects, properties, and laws -- must be integrated into a coherent whole. 
Once a theory has been constructed, it can be applied to predict future sensor readings, retrodict earlier readings, or impute missing readings.

In this paper, we evaluate the \sys{} in a diverse variety of domains, including cellular automata, rhythms and simple nursery tunes, multi-modal binding problems, occlusion tasks, and sequence induction intelligence tests. 
In each domain, we test our engine's ability to predict future sensor values, retrodict earlier sensor values, and impute missing sensory data.
The engine performs well in all these domains, significantly outperforming neural net baselines and state of the art inductive logic programming systems.
These results are significant because neural nets typically struggle to solve the binding problem (where information from different modalities must somehow be combined together into different aspects of one unified object) and fail to solve occlusion tasks (in which objects are sometimes visible and sometimes obscured from view).
We note in particular that in the sequence induction intelligence tests, our system achieved human-level performance.
This is notable because our system is not a bespoke system designed specifically to solve intelligence tests, but a \emph{general-purpose} system that was designed to make sense of \emph{any} sensory sequence.

\end{abstract}


\section{Introduction}
\label{sec:intro}

In a previous paper \cite{evans2019making}, we described the \sys{}\footnote{``Apperception'' comes from the French `apercevoir' \cite{leibniz1996leibniz}. 
Apperception, as we use it in this paper, is the process of assimilating sensory information into a coherent unified whole \cite{dewey1888leibniz}.}, an unsupervised learning system for making sense of sequences of sensory input.

The \sys{} makes sense of sensory input by constructing a symbolic theory that explains the sensory input \cite{tenenbaum2006theory,pasula2007learning,ray2009nonmonotonic,inoue2014learning}.
Following Spelke and others \cite{spelke2007core,diuk2008object}, we assume the theory is composed of objects that persist over time, with properties that change over time according to general laws.
Following John McCarthy and others \cite{mccarthy2006challenges,inoue2016meta,teijeiro2018adoption}, we assume the theory makes sense of the sensory sequence by positing \emph{latent objects}: some sensory sequences can only be made intelligible by hypothesizing an underlying reality, distinct from the surface features of our sensors, that makes the surface phenomena intelligible. 
Once it has constructed such a theory, the engine can apply it to predict future sensor readings, to retrodict past readings, or to impute missing values.

In \cite{evans2019making}, we argued that although constructing a symbolic theory that explains the sensory sequence is necessary for making sense of the sequence, it is not, on its own, sufficient.
There is one further additional ingredient needed to make sense of the sequence.
This is the requirement that our theory exhibits a particular form of \emph{unity}:
the constituents of our theory -- objects, properties, and atoms -- must be integrated into a coherent whole. 
Specifically, our unity condition requires that the objects are interrelated via chains of binary relations, the properties are connected via exclusion relations, and the atoms are unified by jointly satisfying the theory's constraints.
This extra unity condition is necessary, we argued, for the theory to achieve good accuracy at prediction, retrodiction, and imputation.

The \sys{} has a number of appealing features.
(1) Because the causal theories it generates are symbolic, they are human-readable and hence verifiable. 
We can understand precisely how the system is making sense of its sensory data\footnote{Human readability is a much touted feature of Inductive Logic Programming (ILP) systems, but when the learned programs become
large and include a number of invented auxiliary predicates, the resulting programs become less readable
(see \cite{muggleton2018ultra}). But even a large and complex machine-generated logic program will be easier to understand
than a large tensor of floating point numbers.}.
(2) Because of the strong inductive bias (both in terms of the design of the causal language, \logic{}, but also in terms of the unity conditions that must be satisfied), the system is data-efficient, able to make sense of the shortest and scantiest of sensory sequences\footnote{Our sensory sequences are less than 300 bits. See Table \ref{table:experiments-overview}.}.
(3) Our system generates a causal model that is able to accurately predict future sensory input.
But \emph{that is not all it can do}; it is also able to retrodict previous values and impute missing sensory values in the middle of the sensory stream.
In fact, our system is able to predict, retrodict, and impute simultaneously\footnote{See Example \ref{example:one} for a case where the \sys{} jointly predicts, retrodicts, and imputes.}.

In this paper, we evaluate the \sys{} in a diverse variety of domains, with encouraging results. 
The five domains we use are elementary cellular automata, rhythms and nursery tunes, ``Seek Whence'' and C-test sequence induction intelligence tests \cite{hofstadter2008fluid}, multi-modal binding tasks, and occlusion problems. 
These tasks were chosen because they require cognition rather than mere classificatory perception, and because they are simple for humans but not for modern machine learning systems e.g. neural networks\footnote{Figure \ref{fig:baselines-chart} shows how neural baselines struggle to solve these tasks.}. 
The \sys{} performs well in all these domains, significantly out-performing neural net baselines.
These results are significant because neural systems typically struggle to solve the binding problem (where information from different modalities must somehow be combined into different aspects of one unified object) and fail to solve occlusion tasks (in which objects are sometimes visible and sometimes obscured from view).

We note in particular that in the sequence induction intelligence tests, our system achieved human-level performance.
This is notable because the \sys{} was not designed to solve these induction tasks; it is not a bespoke hand-engineered solution to this particular domain. 
Rather, it is a \emph{general-purpose}\footnote{This claim needs to be qualified. Although the algorithm is general-purpose (the exact same code is applied to many different domains), a small amount of domain-specific knowledge is provided for each domain. In particular, in the sequence induction task, the successor relation on letters is given to the system.} system that attempts to make sense of \emph{any} sensory sequence.
This is, we believe, a highly suggestive result \cite{hernandez2016computer}.

In ablation tests, we tested what happened when each of the four unity conditions was turned off.
Since the system's performance deteriorates noticeably when each unity condition is ablated, this indicates that the unity conditions are indeed doing vital work in our engine's attempts to make sense of the incoming barrage of sensory data.


\subsection{Related work}
\label{sec:intro-related}

A human being who has built a mental model of the world can use that model for counterfactual reasoning, anticipation, and planning \cite{craik1967nature,harris2000work,gerstenberg2017intuitive}.
Similarly, computer agents endowed with mental models are able to achieve impressive performance in a variety of domains.
For instance, Lukasz Kaiser et al.~\cite{kaiser2019} show that a model-based RL agent trained on 100K interactions compares with a state-of-the-art model-free agent trained on tens or hundreds of millions of interactions. 
David Silver et al.~\cite{silver2018general} have shown that a model-based Monte Carlo tree search planner with policy distillation can achieve superhuman level performance in a number of board games. The tree search relies, crucially, on an accurate model of the game dynamics.

When we have an accurate model of the environment, we can leverage that model to anticipate and plan.
But in many domains, we do not have an accurate model.
If we want to apply model-based methods in these domains, we must \emph{learn} a model from the stream of observations.
In the rest of this section, we shall describe various different approaches to representing and learning models, 
and show where our particular approach fits into the landscape of model learning systems.

Before we start to build a model to explain a sensory sequence, one fundamental question is: what form should the model take?
We shall distinguish three dimensions of variation of models (adapted from \cite{hamrick2019analogues}): first, whether they simply model the observed phenomena, or whether they also model latent structure; second, whether the model is explicit and symbolic or implicit; and third, what type of prior knowledge is built into the model structure. 

We shall use the hidden Markov model (HMM)\footnote{Many systems predict state dynamics for partially observable Markov decision processes (POMDPs), rather than HMMs. In a POMDP, the state transition function depends on the previous state $z_t$ and the action $a_t$ performed by an agent. See Jessica Hamrick's paper for an excellent overview \cite{hamrick2019analogues} of model-based methods in deep learning that is framed in terms of POMDPs. In this paper, we consider HMMs. Adding actions to our model is not particularly difficult, but is left for further work.} \cite{baum1966statistical,ghahramani2001introduction} as a general framework for describing sequential processes.
Here, the observation at time $t$ is $x_t$, and the latent state is $z_t$.
In a HMM, the observation $x_t$ at time $t$ depends only on the latent (unobserved) state $z_t$. 
The state $z_t$ in turn depends only on the previous latent state $z_{t-1}$.

%
%

The first dimension of variation amongst models is whether they actually use latent state information $z_t$ to explain the observation $x_t$. 
Some approaches \cite{feinberg2018model,nagabandi2018neural,battaglia2016interaction,chang2016compositional,mrowca2018flexible,sanchez2018graph} assume we are \emph{given} the underlying state information $z_{1:t}$.
In these approaches, there is no distinction between the observed phenomena and the latent state: $x_i = z_i$.
With this simplifying assumption, the only thing a model needs to learn is the transition function.
Other approaches \cite{lerer2016learning,finn2017deep,bhattacharyya2018long} focus only on the observed phenomena $x_{1:t}$ and ignore latent information $z_{1:t}$ altogether. 
These approaches predict observation $x_{t+1}$ given observation $x_t$ without positing any hidden latent structure.
Some approaches take latent information seriously \cite{oh2015action,chiappa2017recurrent,ha2018recurrent,buesing2018learning,janner2018reasoning}.
These jointly learn a perception function (that produces a latent $z_t$ from an observed $x_t$), a transition function (producing a next latent state $z_{t+1}$ from latent state $z_t$) and a rendering function (producing a predicted observation $x_{t+1}$ from the latent state $z_{t+1}$).
Our approach also builds a latent representation of the state.
As well as positing latent properties (unobserved properties that explain observed phenomena), 
we also posit latent \emph{objects} (unobserved objects whose relations to observed objects explain observed phenomena).

But our use of latent information is rather different from its use in \cite{oh2015action,chiappa2017recurrent,ha2018recurrent,buesing2018learning,janner2018reasoning}.
In their work, the latent information is merely a lower-dimensional representation of the surface information:
since a neural network represents a function mapping the given sensor information to a latent representation, the latent representation is nothing more than a summary,  a distillation, of the sensory given.
But we use latent information rather differently. Our latent information goes \emph{beyond} the given sensory information to include invented objects and properties that are not observed but \emph{constructed} in order to make sense of what is observed.
Following John McCarthy \cite{mccarthy2006challenges}, we assume that making sense of the surface sensory perturbations requires hypothesizing an underlying reality, distinct from the surface features of our sensors, that makes the surface phenomena intelligible. 

The second dimension of variation concerns whether the learned model is explicit, symbolic and human-readable, or implicit and inscrutable. In some approaches \cite{oh2015action,chiappa2017recurrent,ha2018recurrent,buesing2018learning}, the latent states are represented by vectors and the dynamics of the model by weight tensors. 
In these cases, it is hard to understand what the system has learned. 
In other approaches \cite{zhang2018composable,xu2019,asai2018classical,asai2019unsupervised}, the latent state is represented symbolically, but the state transition function is represented by the weight tensor of a neural network and is inscrutable. We may have some understanding of what state the machine thinks it is in, but we do not understand why it thinks there is a transition from this state to that. In some approaches \cite{ray2009nonmonotonic,inoue2014learning,katzouris2015incremental,michelioudakis2016mathtt,katzouris2016online,michelioudakis2018semi}, both the latent state and the state transition function are represented symbolically.
Here, the latent state is a set of ground atoms\footnote{A ground atom is a logical atom that contains no variables.} and the state transition function is represented by a set of universally quantified rules.
Our approach falls into this third category. 
Here, the model is fully interpretable: we can interpret the state the machine thinks it is in, and we can understand the reason why it believes it will transition to the next state.

A third dimension of variation between models is the amount and type of prior knowledge that they include.
Some model learning systems have very little prior knowledge.
In some of the neural systems (e.g. \cite{finn2017deep}), the only prior knowledge is the spatial invariance assumption implicit in the convolutional network's structure.
Other models incorporate prior knowledge about the way objects and states should be represented.
For example, some models assume objects can be composed in hierarchical structures \cite{xu2019}.
Other systems additionally incorporate prior knowledge about the type of rules that are used to define the state transition function. 
For example, some \cite{michelioudakis2016mathtt,katzouris2016online,michelioudakis2018semi} use prior knowledge of the event calculus \cite{kowalski1989logic}.
Our approach falls into this third category.
We impose a language bias in the form of rules used to define the state transition function and also impose additional requirements on candidate sets of rules: they must satisfy the four unity conditions introduced above (and elaborated in Section \ref{sec:unity-conditions} below).

To summarize, in order to position our approach within the landscape of other approaches, we have distinguished three dimensions of variation. 
Our approach differs from neural approaches in that the posited theory is explicit and human readable. Not only is the representation of state explicit (represented as a set of ground atoms) but the transition dynamics of the system are also explicit (represented as universally quantified rules in a domain specific language designed for describing causal structures).
Our approach differs from other inductive program synthesis methods in that it posits significant latent structure in addition to the induced rules to explain the observed phenomena: in our approach, explaining a sensory sequence does not just mean constructing a set of rules that explain the transitions; it also involves positing a type signature containing a set of latent properties and a set of latent \emph{objects}. 
Our approach also differs from other inductive program synthesis methods in the type of prior knowledge that is used: as well as providing a strong language bias by using a particular representation language (a typed extension of datalog with causal rules and constraints), we also inject a substantial inductive bias: the unity conditions, the key constraints on our system, represent domain-\emph{independent} prior knowledge. 

\subsection{Paper outline}

Section \ref{sec:apperception-framework} provides an introduction to the \sys{}.
Section \ref{sec:experiments} describes our experiments in five different types of task: elementary cellular automata, rhythms and nursery tunes, ``Seek Whence'' sequence induction tasks, multi-modal binding tasks, and occlusion problems.
In Section \ref{sec:results}, we compare our system to neural network baselines, and to a state of the art inductive logic programming system.

\section{Making sense of sensory sequences}
\label{sec:apperception-framework}

In this section, we provide an introduction to the \sys{}.
For a full description, see \cite{evans2019making}.

We use basic concepts and standard notation from logic programming \cite{kowalski1974predicate,apt1990logic,lloyd2012foundations}.
A function-free \define{atom} is an expression of the form $p(t_1,..., t_n)$, where $p$ is a predicate of arity $n \geq 0$ and
each $t_i$ is either a variable or a constant.
A \define{ground atom} is an atom that contains no variables, while an \define{unground atom} is an atom that contains no constants.
We shall use $a, b, c, ...$ for constants, $X, Y, Z, ...$ for variables, and $p, q, r, ...$ for predicate symbols.

A \define{Datalog clause} is a definite clause of the form $\alpha_1 \wedge ... \wedge \alpha_n \rightarrow \alpha_{0}$
where each $\alpha_i$ is an atom and $n \geq 0$. 
A \define{Datalog program} is a set of Datalog clauses.

The \sys{} takes a sequence of sets of sensor readings, and produces a theory that ``makes sense'' of that sequence.
We assume that the sensor readings have already been discretized into ground atoms, so a sensory reading featuring sensor $a$ can be represented by a ground atom $p(a)$ for some unary predicate $p$, or by an atom $r(a,b)$ for some binary relation $r$ and unique value $b$.\footnote{We restrict our attention to unary and binary predicates. This restriction can be made without loss of generality, since every $k$-ary relationship can be expressed as $k+1$ binary relationships \cite{kowalski1979logic}.}

An \define{unambiguous symbolic sensory sequence} is a sequence of sets of ground atoms.
Given a sequence $S = (S_1, S_2, ...)$, every \define{state} $S_t$ is a set of ground atoms, representing a \emph{partial} description of the world at a discrete time step $t$.
An atom $p(a) \in S_t$ represents that sensor $a$ has property $p$ at time $t$.
An atom $r(a, b) \in S_t$ represents that sensor $a$ is related via relation $r$ to value $b$ at time $t$.

The central idea is to make sense of a sensory sequence by \emph{constructing a unified theory that explains that sequence}. The key notions, here, are ``theory'', ``explains'', and ``unified''. We consider each in turn.

\subsection{The theory}
\label{sec:theory}

Theories are defined in a new language, \logic{}, designed for modelling dynamics.
In this language, one can describe how facts change over time by writing a causal rule stating that if the antecedent holds at the current time-step, then the consequent holds at the \emph{next} time-step.
Additionally, our language includes a frame axiom allowing facts to persist over time: each atom remains true at the next time-step unless it is overridden by a new fact which is incompatible with it.
Two facts are incompatible if there is a \emph{constraint} that precludes them from both being true. Thus, \logic{} extends Datalog with causal rules and constraints.

A \define{theory} is a four-tuple $(\phi, I, R, C)$ of \logic{} elements where:
\begin{itemize}
\item
$\phi$ is a type signature specifying the types of constants, variables, and arguments of predicates
\item
$I$ is a set of initial conditions
\item
$R$ is a set of rules describing the dynamics
\item
$C$ is a set of constraints
\end{itemize}

The \define{initial conditions} $I$ is a set of ground atoms representing a partial description of the facts true at the initial time step.
The initial conditions are needed to specify the initial values of the latent unobserved information. 
Some systems (e.g. LFIT \cite{inoue2014learning}) define a predictive model without using a set of initial conditions. These systems are able to avoid positing initial conditions because they do not use latent unobserved information. But any system that does invoke latent information beneath the surface of the sensory stimulations must also define the initial values of the latent information.

The rules $R$ define the dynamics of the theory.
There are two types of rule in $\logic{}$.
A \define{static rule} is a definite clause of the form $\alpha_1 \wedge ... \wedge \alpha_n \rightarrow \alpha_0$, where $n \geq 0$ and each $\alpha_i$ is an \emph{unground} atom consisting of a predicate and a list of variables.
Informally, a static rule is interpreted as: if conditions $\alpha_1, ... \alpha_n$ hold at the current time step, then $\alpha_0$ also holds at that time step.
A \define{causal rule} is a clause of the form $\alpha_1 \wedge ... \wedge \alpha_n \fork \alpha_0$, where $n \geq 0$ and each $\alpha_i$ is an unground atom.
A causal rule expresses how facts change over time. 
Rule $\alpha_1 \wedge ... \wedge \alpha_n \fork \alpha_0$ states that if conditions $\alpha_1, ... \alpha_n$ hold at the current time step, then $\alpha_0$ holds at the \emph{next} time step.

All variables in rules are implicitly universally quantified.
So, for example, $\mathit{on}(X) \fork \mathit{off}(X)$ states that for all objects $X$, if $X$ is currently $\mathit{on}$, then $X$ will become  $\mathit{off}$ at the next-time step.

The constraints $C$ rule out certain combinations of atoms\footnote{Note that exclusive disjunction between atoms $p_1(X), ..., p_n(X)$ is different from \emph{xor} between the $n$ atoms. 
The \emph{xor} of $n$ atoms is true if an \emph{odd} number of the atoms hold, while the exclusive disjunction is true if \emph{exactly one} of the atoms holds.
We write $p_1(X) \oplus ... \oplus p_n(X)$ to mean exclusive disjunction between $n$ atoms, not the application of $n-1$ \emph{xor} operations.}:
There are three types of constraint in \logic{}.
A \define{unary constraint} is an expression of the form
$\forall X, p_1(X) \oplus ... \oplus p_n(X)$, where $n > 1$, meaning that for all $X$, exactly one of $p_1(X), ..., p_n(X)$ holds.
A \define{binary constraint} is an expression of the form
$\forall X, \forall Y, r_1(X, Y) \oplus ... \oplus r_n(X, Y)$ where $n > 1$, meaning that for all objects $X$ and $Y$, exactly one of the binary relations hold.
A \define{uniqueness constraint} is an expression of the form
$\forall X, \exists ! Y{:}t_2, r(X, Y)$,
which means that for all objects $X$ of type $t_1$ there exists a \emph{unique} object $Y$ such that $r(X, Y)$.

Note that the rules and constraints are constructed entirely from \emph{unground} atoms. 
Disallowing constants prevents special-case rules that apply to particular objects, and forces the theory to be general.\footnote{This restriction also occurs in some ILP systems \cite{inoue2014learning,evans2018learning}.}


\subsection{Explaining the sensory sequence}
\label{sec:explaining}

A theory explains a sensory sequence if the theory generates a trace\footnote{In \cite{inoue2014learning}, the trace is called the \emph{orbit}.} that covers that sequence.
In this section, we explain the trace and the covering relation.

Every theory $\theta = (\phi, I, R, C)$ generates an infinite sequence $\tau(\theta)$ of sets of ground atoms, called the \define{trace} of that theory.
Here, $\tau(\theta) = (A_1, A_2, ...)$, where each $A_t$ is the smallest set of atoms satisfying the following conditions:
\begin{itemize}
\item 
$I \subseteq A_1$ 
\item
Each $A_t$ is closed under every static rule in $R$
\item
Each pair $(A_t, A_{t+1})$ of consecutive states is closed under every dynamic rule in $R$
\item
\emph{Frame axiom}: if $\alpha$ is in $A_{t-1}$ and there is no atom in $A_t$ that is incompossible with $\alpha$ w.r.t constraints $C$, then $\alpha \in A_t$. 
Two ground atoms are \define{incompossible} if there is some constraint $c$ in $C$ that precludes both atoms being true.
\end{itemize}
The frame axiom is a simple way of providing inertia: a proposition continues to remain true until something new comes along which is incompatible with it.
Including the frame axiom makes our theories much more concise: instead of needing rules to specify all the atoms which remain the same, we only need rules that specify the atoms that change. 
Note that the state transition function is deterministic: $A_{t}$ is uniquely determined by $A_{t-1}$.

A theory $\theta$ \define{explains} a sensory sequence $(S_1, ..., S_T)$ if the trace of $\theta$ covers $S$. 
The trace $\tau(\theta) = (A_1, A_2, ...)$ covers $(S_1, ..., S_T)$ if each $S_i$ in the sensory sequence is a subset of the corresponding $A_i$ in the trace.

In providing a theory $\theta$ that explains a sensory sequence $S$, we make $S$ intelligible by placing it within a bigger picture:
while $S$ is a scanty and incomplete description of a fragment of the time-series, $\tau(\theta)$ is a complete and determinate description of the whole time-series.

\begin{example}
\label{ex:sensory-sequence}
Consider, the following sequence $S_{1:10}$.
Here there are two sensors $a$ and $b$, and each sensor can be either $\mathit{on}$ or $\mathit{off}$.
\begin{eqnarray*}
\begin{tabular}{lllll}
$S_1 = \set{}$  & $S_2 = \set{ \mathit{off}(a), \mathit{on}(b)}$ & $S_3 = \set{ \mathit{on}(a),  \mathit{off}(b)}$ & 
$S_4 = \set{ \mathit{on}(a), \mathit{on}(b)}$ & $S_5 = \set{ \mathit{on}(b)}$ \\ 
$S_6 = \set{ \mathit{on}(a),  \mathit{off}(b)}$ & $S_7 = \set{ \mathit{on}(a), \mathit{on}(b)}$ & $S_8 = \set{ \mathit{off}(a), \mathit{on}(b)}$ & $S_9 = \set{ \mathit{on}(a)}$ & $S_{10} = \set{ }$
\end{tabular}
\end{eqnarray*}
There is no expectation that a sensory sequence contains readings for all sensors at all time steps.
Some of the readings may be missing.
In state $S_5$, we are missing a reading for $a$, while in state $S_9$, we are missing a reading for $b$.
In states $S_1$ and $S_{10}$, we are missing sensor readings for both $a$ and $b$.

Consider the type signature $\phi = (T, O, P, V)$, consisting of types $T = \set{s}$, objects $O = \set{a{:}s, b{:}s}$, predicates $P = \set{ \mathit{on}(s), \mathit{off}(s), p_1(s), p_2(s), p_3(s), r(s, s)}$, and variables $V=\set{X {:}s, Y {:}s}$.
Consider the theory $\theta = (\phi, I, R, C)$, where:
\begin{eqnarray*}
\begin{tabular}{lll}
$I = \bigsetbegin{}
p_1(b) \\
p_2(a) \\
r(a, b) \\
r(b, a) \\
\bigsetend{}$ &
$R = \bigsetbegin{}
p_1(X) \fork p_2(X) \\
p_2(X) \fork p_3(X) \\
p_3(X) \fork p_1(X) \\
p_1(X) \rightarrow \mathit{on}(X) \\
p_2(X) \rightarrow \mathit{on}(X) \\
p_3(X) \rightarrow \mathit{off}(X)
\bigsetend{}$ &
$C = \bigsetbegin{}
\forall X {:}s, \; \mathit{on}(X) \oplus \mathit{off}(X) \\
\forall X {:}s, \; p_1(X) \oplus p_2(X) \oplus p_3(X) \\
\forall X {:}s, \; \exists ! Y {:}s \; r(X, Y)
\bigsetend{}$
\end{tabular}
\end{eqnarray*}

The infinite trace $\tau(\theta) = (A_1, A_2, ...)$ begins with:
\begin{eqnarray*}
\begin{tabular}{ll}
$A_1 = \{ \mathit{on}(a), \mathit{on}(b), p_2(a), p_1(b), r(a, b), r(b, a) \}$ &
$A_2 = \{ \mathit{off}(a), \mathit{on}(b), p_3(a), p_2(b), r(a, b), r(b, a) \}$ \\
$A_3 = \{ \mathit{on}(a), \mathit{off}(b), p_1(a), p_3(b), r(a, b), r(b, a) \}$ &
$A_4 = \{ \mathit{on}(a), \mathit{on}(b), p_2(a), p_1(b), r(a, b), r(b, a) \}$ \\ 
\dotso 
\end{tabular}
\end{eqnarray*}
Note that the trace repeats at step 4. In fact, it is always true that the trace repeats after some finite set of time steps.

The theory $\theta$ explains the sensory sequence $S$, since the trace $\tau(\theta)$ covers $S$.
Note that $\tau(\theta)$ ``fills in the blanks'' in the original sequence $S$, both predicting final time step 10, retrodicting   initial time step 1, and imputing missing values for time steps 5 and 9.

\label{example:one}
\end{example}

\subsection{Unifying the sensory sequence}
\label{sec:unity-conditions}

In order for theory $\theta$ to make sense of sequence $S$, it is \emph{necessary} that $\tau(\theta)$ covers $S$.
But this condition is not, on its own, \emph{sufficient}.
The extra condition that is needed for $\theta$ to count as ``making sense'' of $S$ is for $\theta$ to be \emph{unified}.

We require that the constituents of the theory are integrated into a coherent whole.
A trace $\tau(\theta)$ of theory $\theta$ is a (i) sequence of (ii) sets of ground atoms composed of (iii) predicates and (iv) objects.
For the theory $\theta$ to be unified is for unity to be achieved at each of these four levels.

A theory $\theta = (\phi, I, R, C)$ is \define{unified} if each of the four conditions hold:
\begin{enumerate}
\item
Objects are united in space: for each state $A_t$ in $\tau(\theta) = (A_1, A_2, ...)$, for each pair $(x, y)$ of distinct objects, $x$ and $y$ are connected via a chain of binary atoms $\{r_1(x, z_1), r_2(z_1, z_2), ... r_n(z_{n-1}, z_n), r_{n+1}(z_n, y)\} \subseteq A_t$.
\item
Predicates are united via constraints: for each unary predicate $p$ in $\phi$, there is some xor constraint in $C$ of the form $\forall X{:}t, p(X) \oplus q(X) \oplus ...$ containing $p$;
and, for each binary predicate $r$ in $\phi$, there is some xor constraint in $C$ of the form $\forall X{:}t_1, \forall Y{:}t_2, r(X, Y) \oplus s(X, Y) \oplus ...$ or some $\exists!$ constraint in $C$ of the form $\forall X{:}t, \exists ! Y{:}t_2, r(X, Y)$.
\item
Ground atoms are united into states by jointly respecting constraints: each constraint in $C$ is satisfied in each state of the trace.
\item
States are united into a sequence by causal rules: each consecutive pair of states $(A_t, A_{t+1})$ in the trace satisfies every causal rule in $R$.
\end{enumerate}

Now we are ready to define the central notion of ``making sense'' of a sequence.
A theory $\theta$ \define{makes sense} of a sensory sequence $S$ if the trace of $\theta$ covers $S$ and $\theta$ satisfies the four conditions of unity.
If $\theta$ makes sense of $S$, we also say that $\theta$ is a \define{unified interpretation} of $S$.

\subsection{Examples}

In this section, we provide a worked example of an apperception task, along with different unified interpretations.
We wish to highlight that there are always many alternative ways of interpreting a sensory sequence, each with different latent information (although some may have higher cost than others).

We continue to use our running example, the sensory sequence from Example \ref{example:one}.
Here there are two sensors $a$ and $b$, and each sensor can be $\mathit{on}$ or $\mathit{off}$.
\begin{eqnarray*}
\begin{tabular}{lllll}
$S_1 = \set{}$  & $S_2 = \set{ \mathit{off}(a), \mathit{on}(b)}$ & $S_3 = \set{ \mathit{on}(a),  \mathit{off}(b)}$ & 
$S_4 = \set{ \mathit{on}(a), \mathit{on}(b)}$ & $S_5 = \set{ \mathit{on}(b)}$ \\ 
$S_6 = \set{ \mathit{on}(a),  \mathit{off}(b)}$ & $S_7 = \set{ \mathit{on}(a), \mathit{on}(b)}$ & $S_8 = \set{ \mathit{off}(a), \mathit{on}(b)}$ & $S_9 = \set{ \mathit{on}(a)}$ & $S_{10} = \set{ }$
\end{tabular}
\end{eqnarray*}
Let $\phi = (T, O, P, V)$ where $T = \{\mathit{sensor}\}$, $O = \{a, b\}$, $P = \{\mathit{on}(\mathit{sensor}), \mathit{off}(\mathit{sensor})\}$, $V = \{X {:}\mathit{sensor}\}$.

Examples \ref{example:eca1}, \ref{example:eca2}, and \ref{example:eca3} below show three different unified interpretations of Example \ref{example:one}.

\begin{example}
\label{example:eca1}
One possible way of interpreting Example \ref{example:one} is as follows.
The sensors $a$ and $b$ are simple state machines that cycle between states $p_1$, $p_2$, and $p_3$.
Each sensor switches between $\mathit{on}$ and $\mathit{off}$ depending on which state it is in. 
When it is in states $p_1$ or $p_2$, the sensor is on; when it is in state $p_3$, the sensor is off.
In this interpretation, the two state machines $a$ and $b$ do not interact with each other in any way.
Both sensors are following the same state transitions.
The reason the sensors are out of sync is because they start in different states.

The type signature for this first unified interpretation is $\phi' = (T, O, P, V)$, where $T = \set{\mathit{sensor}}$, 
$O = \set{a{:}\mathit{sensor}, b{:}\mathit{sensor}}$, 
$P = \set{ \mathit{on}(\mathit{sensor}), \mathit{off}(\mathit{sensor}), r(\mathit{sensor}, \mathit{sensor}), p_1(\mathit{sensor}), p_2(\mathit{sensor}), p_3(\mathit{sensor})}$, and 
$V = \set{X {:}\mathit{sensor}, Y {:}\mathit{sensor}}$.
The three unary predicates $p_1$, $p_2$, and $p_3$ are used to represent the three states of the state machine.

Our first unified interpretation is the tuple $(\phi', I, R, C')$, where:
\begin{eqnarray*}
\begin{tabular}{lll}
$I = \bigsetbegin{}
p_2(a) \\
p_1(b) \\
r(a, b) \\
r(b, a) \\
\bigsetend{}$ &
$R = \bigsetbegin{}
p_1(X) \fork p_2(X) \\
p_2(X) \fork p_3(X) \\
p_3(X) \fork p_1(X) \\
p_1(X) \rightarrow \mathit{on}(X) \\
p_2(X) \rightarrow \mathit{on}(X) \\
p_3(X) \rightarrow \mathit{off}(X)
\bigsetend{}$ &
$C' = \bigsetbegin{}
\forall X {:}\mathit{sensor}, \; \mathit{on}(X) \oplus \mathit{off}(X) \\
\forall X {:}\mathit{sensor}, \; p_1(X) \oplus p_2(X) \oplus p_3(X) \\
\forall X {:}\mathit{sensor}, \; \exists ! Y {:}\mathit{sensor} \; r(X, Y)
\bigsetend{}$
\end{tabular}
\end{eqnarray*}
The update rules $R$ contain three causal rules (using $\fork$) describing how each sensor cycles from state $p_1$ to $p_2$ to $p_3$, and then back again to $p_1$.
For example, the causal rule $p_1(X) \fork p_2(X)$ states that if sensor $X$ satisfies $p_1$ at time $t$, then $X$ satisfies $p_2$ at time $t+1$. We know that $X$ is a sensor from the variable typing information in $\phi'$.
$R$ also contains three static rules (using $\rightarrow$) describing how the $\mathit{on}$ or $\mathit{off}$ attribute of a sensor depends on its state.
For example, the static rule $p_1(Y) \rightarrow \mathit{on}(X) $ states that if $X$ satisfies $p_1$ at time $t$, then $X$ also satisfies $\mathit{on}$ at time $t$.

The constraints $C'$ state that (i) every sensor is (exclusively) either $\mathit{on}$ or $\mathit{off}$, that every sensor is (exclusively) either $p_1$, $p_2$, or $p_3$, and that every sensor has exactly one sensor that is related by $r$ to it.
The binary $r$ predicate, or something like it, is needed to satisfy the constraint of spatial unity.

In this first interpretation, three new predicates are invented ($p_1$, $p_2$, and $p_3$) to represent the three states of the state machine.
In the next interpretation, we will introduce new invented objects instead of invented predicates.

Given the initial conditions $I$ and the update rules $R$, we can use our interpretation to compute which atoms hold at which time step.
In this case, $\tau(\theta) = (A_1, A_2, ...)$ where $S_i \sqsubseteq A_i$.
Note that this trace repeats: $A_i = A_{i+3}$.
We can use the trace to predict the future values of our two sensors at time step 10, since
$A_{10} = \set{\mathit{on}(a), \mathit{on}(b), r(a, b), r(b, a), p_2(a), p_1(b)}$.

As well as being able to predict future values, we can retrodict past values (filling in $A_1$), or interpolate intermediate unknown values (filling in $A_5$ or $A_9$).\footnote{This ability to ``impute'' intermediate unknown values is straightforward given an interpretation. Recent results show that current neural methods for sequence learning are more comfortable predicting future values than imputing intermediate values.}
But although an interpretation provides the resources to ``fill in'' missing data, it has no particular bias to predicting future time-steps. 
The conditions which it is trying to satisfy (the unity conditions of Section \ref{sec:unity-conditions}) do not explicitly insist that an interpretation must be able to predict future time-steps.
Rather, the ability to predict the future (as well as the ability to retrodict the past, or interpolate intermediate values) is a \emph{derived} capacity that emerges from the more fundamental capacity to ``make sense'' of the sensory sequence.

\end{example}

\begin{example}
\label{example:eca2}

There are always infinitely many different ways of interpreting a sensory sequence.
Next, we show a rather different interpretation of $S_{1:10}$ from that of Example \ref{example:eca1}.
In our second unified interpretation, we no longer see sensors $a$ and $b$ as self-contained state-machines.
Now, we see the states of the sensors as depending on their left and right neighbours. 
In this new interpretation, we no longer need the three invented unary predicates ($p_1$, $p_2$, and $p_3$), but instead introduce a new \emph{object}.

Object invention is much less explored than predicate invention in inductive logic programming. 
But Dietterich et al.~\cite{dietterich2008structured} anticipated the need for it, and Inoue \cite{inoue2016meta} uses meta-level abduction to posit unperceived objects.

Our new type signature $\phi' = (T, O, P, V)$, where
$T= \set{\mathit{sensor}}$, 
$O = \set{a{:}\mathit{sensor}, b{:}\mathit{sensor}, c{:}\mathit{sensor}}$, \\
$P = \set{ \mathit{on}(\mathit{sensor}), \mathit{off}(\mathit{sensor}), r(\mathit{sensor}, \mathit{sensor})}$, and 
$V = \set{X{:}\mathit{sensor}, Y{:}\mathit{sensor}}$.

In this new interpretation, imagine there is a one-dimensional cellular automaton with three cells, $a$, $b$, and (unobserved) $c$. 
The three cells wrap around: the right neighbour of $a$ is $b$, the right neighbour of $b$ is $c$, and the right neighbour of $c$ is $a$.
In this interpretation, the spatial relations are fixed. (We shall see another interpretation later where this is not the case).
The cells alternate between on and off according to the following simple rule: if $X$'s left neighbour is on (respectively off) at $t$, then $X$ is on (respectively off) at $t+1$.

Note that objects $a$ and $b$ are the two sensors we are given, but $c$ is a new unobserved latent object that we posit in order to make sense of the data. 
Many interpretations follow this pattern: new latent unobserved objects are posited to make sense of the changes to the sensors we are given.

Note further that part of finding an interpretation is constructing the spatial relation between objects; this is not something we are given, but something we must \emph{construct}. In this case, we posit that the imagined cell $c$ is inserted to the right of $b$ and to the left of $a$.

We represent this interpretation by the tuple $(\phi', I, R, C')$, where:
\begin{eqnarray*}
\begin{tabular}{lll}
$I = \bigsetbegin{}
\mathit{on}(a) \\
\mathit{on}(b) \\
\mathit{off}(c) \\
r(a, b) \\
r(b, c) \\
r(c, a)
\bigsetend{}$ & 
$R = \bigsetbegin{}
r(X, Y) \wedge \mathit{off}(X) \fork \mathit{off}(Y) \\
r(X, Y) \wedge \mathit{on}(X) \fork \mathit{on}(Y)
\bigsetend{} $ &
$C' = \bigsetbegin{}
\forall X{:}\mathit{sensor}, \; \mathit{on}(X) \oplus \mathit{off}(X) \\
\forall X{:}\mathit{sensor}, \; \exists ! Y{:}\mathit{sensor}, \; r(X, Y)
\bigsetend{}$
\end{tabular}
\end{eqnarray*}
Here, $\phi'$ extends $\phi$, $C'$ extends $C$, and the interpretation satisfies the unity conditions.
\end{example}

\begin{example}
\label{example:eca3}
We shall give one more way of interpreting the same sensory sequence, to show the variety of possible interpretations.

In our third interpretation, we will posit three latent cells, $c_1$, $c_2$, and $c_3$ that are distinct from the sensors $a$ and $b$.
Cells have static attributes: each cell can be either black or white, and this is a permanent unchanging feature of the cell.
Whether a sensor is on or off depends on whether the cell it is currently contained in is black or white.
The reason why the sensors change from on to off is because they \emph{move} from one cell to another.

Our new type signature $(T, O, P, V)$ distinguishes between cells and sensors as separate types:
$T = \set{\mathit{cell}, \mathit{sensor}}$, 
$O = \set{a: \mathit{sensor}, b: \mathit{sensor}, c_1: \mathit{cell}, c_2: \mathit{cell}, c_3: \mathit{cell}}$, 
$P = \set{ \mathit{on}(\mathit{sensor}), \mathit{off}(\mathit{sensor}), \mathit{part}(\mathit{sensor}, \mathit{cell}), r(\mathit{cell}, \mathit{cell}), \mathit{black}(\mathit{cell}), \mathit{white}(\mathit{cell})}$, and
$V = \set{X: \mathit{sensor}, Y: \mathit{cell}, Y_2: \mathit{cell}}$.
Our interpretation is the tuple $(\phi, I, R, C)$, where:
\begin{eqnarray*}
\begin{tabular}{lll}
$I = \bigsetbegin{}
\mathit{part}(a, c_1) \\
\mathit{part}(b, c_2) \\
r(c_1, c_2) \\
r(c_2, c_3) \\
r(c_3, c_1) \\
\mathit{black}(c_1) \\
\mathit{black}(c_2) \\
\mathit{white}(c_3)
\bigsetend{}$ &
$R = \bigsetbegin{}
\mathit{part}(X,Y) \wedge \mathit{black}(Y) \rightarrow \mathit{on}(X) \\
\mathit{part}(X,Y) \wedge \mathit{white}(Y) \rightarrow \mathit{off}(X) \\
r(Y,Y_2) \wedge \mathit{part}(X,Y_2) \fork \mathit{part}(X,Y)
\bigsetend{}$ &
$C = \bigsetbegin{}
\forall X{:}\mathit{sensor}, \; \mathit{on}(X) \oplus \mathit{off}(X) \\
\forall Y{:}\mathit{cell}, \; \mathit{black}(Y) \oplus \mathit{white}(Y) \\
\forall X{:}\mathit{sensor}, \; \exists ! Y: \mathit{cell}, \; \mathit{part}(X, Y) \\
\forall Y{:}\mathit{cell}, \; \exists ! Y_2: \mathit{cell}, \; r(Y, Y_2)
\bigsetend{}$
\end{tabular}
\end{eqnarray*}
The update rules $R$ state that the $\mathit{on}$ or $\mathit{off}$ attribute of a sensor depends on whether its current cell is black or white.
They also state that the sensors move from right-to-left through the cells.

In this interpretation, there is no state information in the sensors. 
All the variability is explained by the sensors moving from one static object to another.

Here, the sensors move about, so spatial unity is satisfied by different sets of atoms at different time-steps.
For example, at time-step 1, sensors $a$ and $b$ are indirectly connected via the ground atoms $\mathit{part}(a, c_1), r(c_1, c_2), \mathit{part}(b, c_2)$.
But at time-step 2, $a$ and $b$ are indirectly connected via a different set of ground atoms $\mathit{part}(a, c_3), r(c_3, c_1), \mathit{part}(b, c_1)$.
Spatial unity requires all pairs of objects to always be connected via some chain of ground atoms at each time-step, but it does not insist that it is the \emph{same} set of ground atoms at each time-step.
\end{example}

Examples \ref{example:eca1}, \ref{example:eca2}, and \ref{example:eca3} provide different ways of interpreting the same sensory input.
In Example \ref{example:eca1}, the sensors are interpreted as self-contained state machines. 
Here, there are no causal interactions between the sensors: each is an isolated machine, a Leibnitzian monad.
In Examples \ref{example:eca2} and \ref{example:eca3}, by contrast, there are causal interactions between the sensors.
In Example \ref{example:eca2}, the $\mathit{on}$  and $\mathit{off}$ attributes move from left to right along the sensors.
In Example \ref{example:eca3}, it is the sensors that move, not the attributes, moving from right to left.
The difference between these two interpretations is in terms of what is moving and what is static.

Note that the interpretations of Examples \ref{example:eca1}, \ref{example:eca2}, and \ref{example:eca3} have costs 16, 12, and 17 respectively.
So the theory of Example \ref{example:eca2}, which invents an unseen object, is preferred to the other theories that posit more complex dynamics.

\subsection{Computer implementation}
\label{sec:system}

The \sys{} is our system for solving apperception tasks.\footnote{The source code is available at https://github.com/RichardEvans/apperception.}
Given as input a sensory sequence $S$, the engine searches for the smallest theory $\theta = (\phi, I, R, C)$ that makes sense of $S$.
In this section, we sketch how this is implemented.

A \define{template} is a structure for circumscribing a large but finite set of theories. It is a type signature together with constants that bound the complexity of the rules in the theory.
Formally, a template $\chi$ is a tuple $(\phi, N_{\rightarrow}, N_{\fork}, N_B)$ where $\phi$ is a type signature, $N_{\rightarrow}$ is the max number of static rules allowed in $R$, $N_{\fork}$ is the max number of causal rules allowed in $R$,
and $N_B$ is the max number of atoms allowed in the body of a rule in $R$.

Each template specifies a large (but finite) set of theories that conform to it.
Our method is an anytime algorithm that enumerates templates of increasing complexity. 
For each template, it finds the smallest theory $\theta$ that explains the input sequence and satisfies the conditions of unity.
If it finds such a $\theta$, it stores it.
When it has run out of processing time, it returns the lowest cost $\theta$ it has found from all the templates it has considered.

The hardest part of this algorithm is the way we we find the lowest-cost theory $\theta$ for a given template.
Our program synthesis method uses meta-interpretive learning \cite{mugg:metagold,inoue2016meta}: we implement a \logic{} interpreter in Answer Set Programming (ASP) that computes the trace of a theory, we define the covering relation between the sensory sequence and the trace, we define the unity condition as ASP constraints, and use ASP choice rules \cite{gebser2012answer} to choose which initial conditions, rules, and constraints to include in the theory.
This method is described in detail in \cite{evans2019making}.

\section{Experiments}
\label{sec:experiments}

To evaluate the generality of our system, we tested it in a variety of domains: elementary (one-dimensional) cellular automata,
drum rhythms and nursery tunes, sequence induction intelligence tests, multi-modal binding tasks, and occlusion tasks.
These particular domains were chosen because they represent a diverse range of tasks that are simple for humans but are hard for state of the art machine learning systems.
The tasks were chosen to highlight the difference between mere perception (the classification tasks that machine learning systems already excel at) and apperception (assimilating information into a coherent integrated theory, something traditional machine learning systems are not designed to do).
Although all the tasks are data-sparse and designed to be easy for humans but hard for machines, in other respects the domains were chosen to maximize diversity: the various domains involve different sensory modalities, and some sensors provide binary discriminators while others are multi-class.

\subsection{Experimental setup}
\label{sec:experimental-setup}

We implemented the \sys{} in Haskell and ASP.
We used \verb|clingo| \cite{gebser2014clingo} to solve the ASP programs generated by our system.
We ran all experiments with a time-limit of 4 hours on a standard Unix desktop.
We ran \verb|clingo| in ``vanilla'' mode, and did not experiment with the various command-line options for optimization, although it is possible we could achieve significant speedup with judicious use of these parameters.

%

For each of the five domains, we provided an infinite sequence of templates (implemented in Haskell as an infinite list).
Each template sequence is a form of declarative bias \cite{de2012declarative}. 
It is important to note that the domain-specific template sequence is not essential to the \sys{}, as our system can also operate using the domain-\emph{independent} template iterator described in \cite{evans2019making}.
Every template in the template sequence will eventually be found by the domain-independent iterator. 
However, in practice, the \sys{} will find results in a more timely manner when it is given a domain-specific template sequence rather than the domain-independent templates sequence\footnote{This is analogous to the situation in Metagol, which uses \emph{metarules} as a form of declarative bias. As shown in \cite{cropper2015logical}, there is a pair of highly general metarules which are sufficient to entail all metarules of a certain broad class. However, in practice, it is significantly more efficient to use a domain-specific set of metarules, rather than the very general pair of metarules \cite{crop:thesis,cropper2018derivation}.}.

\subsection{Results}

Our experiments (on the prediction task) are summarised in Table \ref{table:experiments-overview}.
Note that our accuracy metric for a single task is rather exacting: the model is accurate (Boolean) on a task iff \emph{every} hidden sensor value is predicted correctly.\footnote{The reason for using this strict notion of accuracy is that, as the domains are deterministic and noise-free, there is a simplest possible theory that explains the sensory sequence. In such cases where there is a correct answer, we wanted to assess whether the system found that correct answer exactly -- not whether it was fortunate enough to come close while misunderstanding the underlying dynamics.} It does not score any points for predicting most of the hidden values correctly.
As can be seen from Table \ref{table:experiments-overview}, our system is able to achieve good accuracy across all five domains.

\begin{table}
\centering
\begin{tabular}{|l|p{1.2cm}|p{1.2cm}|p{1.7cm}|p{2.2cm}|p{1.5cm}|}
\hline
{\bf Domain} & {\bf Tasks (\#)} & {\bf Memory (megs)} & {\bf Input size (bits)} &  {\bf Held out size (bits)} & {\bf Accuracy (\%)} \\
\hline
ECA & 256 & 473.2 & 154.0 & 10.7 & 97.3\% \\
\hline
Rhythm \& music & 30 & 2172.5 & 214.4 & 15.3 & 73.3\% \\
\hline
\emph{Seek Whence} & 30 & 3767.7 & 28.4 & 2.5 & 76.7\% \\
\hline
Multi-modal binding & 20 & 1003.2 & 266.0 & 19.1 & 85.0\% \\
\hline
Occlusion & 20 & 604.3 & 109.2 & 10.1 & 90.0 \% \\
\hline
\end{tabular}
\caption[Results for prediction tasks on the five experimental domains]{Results for prediction tasks on five domains. We show the mean information size of the sensory input, to stress the scantiness of our sensory sequences. 
We also show the mean information size of the held-out data.
Our metric of accuracy for prediction tasks is whether the system predicted \emph{every} sensor's value correctly. }
\label{table:experiments-overview}
\end{table}

In Table \ref{table:kappa-metric}, we display Cohen's kappa coefficient \cite{cohen1960coefficient} for the five domains.
If $a$ is the accuracy and $r$ is the chance of randomly agreeing with the actual classification, then the kappa metric $\kappa = \frac{a - r}{1 - r}$. Since our accuracy metric for a single task is rather exacting (since the model is accurate only if \emph{every} hidden sensor value is predicted correctly), the chance $r$ of random accuracy is very low. 
For example, in the ECA domain with 11 cells, the chance of randomly predicting correctly is $2^{-11}$. 
Similarly, in the music domain, if there are 8 sensors and each can have 4 loudness levels, then the chance of randomly predicting correctly is $4^{-8}$.
Because the chance of random accuracy is so low, the kappa metric closely tracks the accuracy. 

\begin{table}
\centering
\begin{tabular}{|l|p{1.7cm}|p{2.2cm}|p{1.8cm}|}
\hline
{\bf Domain} & {\bf Accuracy $(a)$ } & {\bf Random agreement $(r)$} & {\bf Kappa metric $(\kappa)$} \\
\hline
ECA & 0.973\% & 0.00048 & 0.972 \\
\hline
Rhythm \& music & 0.733\% & 0.00001 & 0.732 \\
\hline
\emph{Seek Whence} & 0.767\% & 0.16666 & 0.720 \\
\hline
Multi-modal binding & 0.850\% & 0.00003 & 0.849 \\
\hline
Occlusion & 0.900 \% & 0.03846 & 0.896  \\
\hline
\end{tabular}
\caption[Cohen's kappa coefficient for the five experimental domains]{Cohen's kappa coefficient for the five domains. Note that the chance of random agreement $(r)$ is very low because we define accuracy as correctly predicting \emph{every} sensor reading. When $r$ is very low, the $\kappa$ metric closely tracks the accuracy.}
\label{table:kappa-metric}
\end{table}

\subsection{Elementary cellular automata}
\label{sec:eca}

An Elementary Cellular Automaton (ECA) \cite{wolfram1983statistical,cook2004universality} is a one-dimensional Cellular Automaton. 
The world is a circular array of cells. 
Each cell can be either $\mathit{on}$ or $\mathit{off}$.
The state of a cell depends only on its previous state and the previous state of its left and right neighbours.

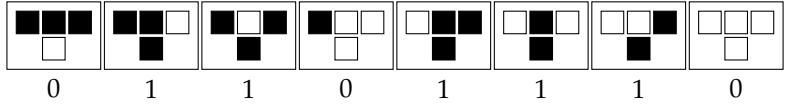
\begin{figure}
\begin{center}
\begin{tikzpicture}[b/.style={draw, minimum size=3mm,   
       fill=black},w/.style={draw, minimum size=3mm},
       m/.style={matrix of nodes, column sep=1pt, row sep=1pt, draw, label=below:#1}, node distance=1pt]

\matrix (A) [m=0]{
|[b]|&|[b]|&|[b]|\\
&|[w]|\\
};
\matrix (B) [m=1, right=of A]{
|[b]|&|[b]|&|[w]|\\
&|[b]|\\
};
\matrix (C) [m=1, right=of B]{
|[b]|&|[w]|&|[b]|\\
&|[b]|\\
};
\matrix (D) [m=0, right=of C]{
|[b]|&|[w]|&|[w]|\\
&|[w]|\\
};
\matrix (E) [m=1, right=of D]{
|[w]|&|[b]|&|[b]|\\
&|[b]|\\
};
\matrix (F) [m=1, right=of E]{
|[w]|&|[b]|&|[w]|\\
&|[b]|\\
};
\matrix (G) [m=1, right=of F]{
|[w]|&|[w]|&|[b]|\\
&|[b]|\\
};
\matrix (H) [m=0, right=of G]{
|[w]|&|[w]|&|[w]|\\
&|[w]|\\
};
\end{tikzpicture}  
\caption[Updates for ECA rule 110]{Updates for ECA rule 110. The top row shows the context: the target cell together with its left and right neighbour. The bottom row shows the new value of the target cell given the context. A cell is black if it is on and white if it is off.}
\label{fig:update110}
\end{center}
\end{figure}
Figure \ref{fig:update110} shows one set of ECA update rules\footnote{This particular set of update rules is known as Rule $110$. Here, $110$ is the decimal representation of the binary $01101110$ update rule, as shown in Figure \ref{fig:update110}.  This rule has been shown to be Turing-complete \cite{cook2004universality}.}. 
Each update specifies the new value of a cell based on its previous left neighbour, its previous value, and its previous right neighbour. 
The top row shows the values of the left neighbour, previous value, and right neighbour.
The bottom row shows the new value of the cell.
There are 8 updates, one for each of the $2^3$ configurations.
In the diagram, the leftmost update states that if the left neighbour is $\mathit{on}$, and the cell is $\mathit{on}$, and its right neighbour is $\mathit{on}$, then at the next time-step, the cell will be turned $\mathit{off}$.
Given that each of the $2^3$ configurations can produce $\mathit{on}$ or $\mathit{off}$ at the next time-step, there are $2^{2^3} = 256$ total sets of update rules.

\begin{figure}
\begin{center}
\begin{tikzpicture}[b/.style={draw, minimum size=3mm,   
       fill=black},w/.style={draw, minimum size=3mm},
       m/.style={matrix of nodes, column sep=1pt, row sep=1pt, draw, label=below:#1}, node distance=1pt]

\matrix (A) [m=110]{
|[w]|&|[w]|&|[w]|&|[w]|&|[w]|&|[b]|&|[w]|&|[w]|&|[w]|&|[w]|&|[w]|\\
|[w]|&|[w]|&|[w]|&|[w]|&|[b]|&|[b]|&|[w]|&|[w]|&|[w]|&|[w]|&|[w]|\\
|[w]|&|[w]|&|[w]|&|[b]|&|[b]|&|[b]|&|[w]|&|[w]|&|[w]|&|[w]|&|[w]|\\
|[w]|&|[w]|&|[b]|&|[b]|&|[w]|&|[b]|&|[w]|&|[w]|&|[w]|&|[w]|&|[w]|\\
|[w]|&|[b]|&|[b]|&|[b]|&|[b]|&|[b]|&|[w]|&|[w]|&|[w]|&|[w]|&|[w]|\\
|[b]|&|[b]|&|[w]|&|[w]|&|[w]|&|[b]|&|[w]|&|[w]|&|[w]|&|[w]|&|[w]|\\
|[b]|&|[b]|&|[w]|&|[w]|&|[b]|&|[b]|&|[w]|&|[w]|&|[w]|&|[w]|&|[b]|\\
|[w]|&|[b]|&|[w]|&|[b]|&|[b]|&|[b]|&|[w]|&|[w]|&|[w]|&|[b]|&|[b]|\\
|[b]|&|[b]|&|[b]|&|[b]|&|[w]|&|[b]|&|[w]|&|[w]|&|[b]|&|[b]|&|[b]|\\
|[w]|&|[w]|&|[w]|&|[b]|&|[b]|&|[b]|&|[w]|&|[b]|&|[b]|&|[w]|&|[w]|\\
|[w]|&|[w]|&|[w]|&|[w]|&|[w]|&|[w]|&|[w]|&|[w]|&|[w]|&|[w]|&|[w]|\\
};
\node at (A-11-1) {?};
\node at (A-11-2) {?};
\node at (A-11-3) {?};
\node at (A-11-4) {?};
\node at (A-11-5) {?};
\node at (A-11-6) {?};
\node at (A-11-7) {?};
\node at (A-11-8) {?};
\node at (A-11-9) {?};
\node at (A-11-10) {?};
\node at (A-11-11) {?};
\end{tikzpicture}  
\end{center}
\caption[One trajectory for ECA rule 110]{One trajectory for Rule 110. Each row represents the state of the ECA at one time-step. In this prediction task, the bottom row (representing the final time-step) is held out. }
\label{fig:trajectory110}
\end{figure}
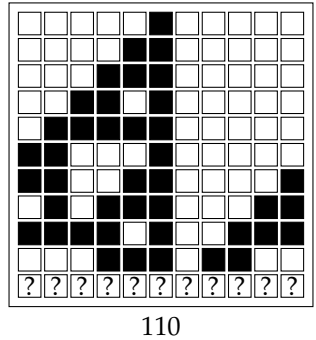

Given update rules for each of the 8 configurations, and an initial starting state, the trajectory of the ECA is determined.
Figure \ref{fig:trajectory110} shows the state sequence for Rule 110 above from one initial starting state of length 11.

In our experiments, we attach sensors to each of the 11 cells, produce a sensory sequence, and then ask our system to find an interpretation that makes sense of the sequence.
For example, for the state sequence of Figure \ref{fig:trajectory110}, the sensory sequence is $(S_1, ..., S_{10})$ where:
\begin{eqnarray*}
S_1 & = & \set{\mathit{off}(c_1), \mathit{off}(c_2), \mathit{off}(c_3), \mathit{off}(c_4), \mathit{off}(c_5), \mathit{on}(c_6), \mathit{off}(c_7), \mathit{off}(c_8), \mathit{off}(c_(), \mathit{off}(c_{10}), \mathit{off}(c_{11})} \\
S_2 & = & \set{\mathit{off}(c_1), \mathit{off}(c_2), \mathit{off}(c_3), \mathit{off}(c_4), \mathit{on}(c_5), \mathit{on}(c_6), \mathit{off}(c_7), \mathit{off}(c_8), \mathit{off}(c_(), \mathit{off}(c_{10}), \mathit{off}(c_{11})} \\
S_3 & = & \set{\mathit{off}(c_1), \mathit{off}(c_2), \mathit{off}(c_3), \mathit{on}(c_4), \mathit{on}(c_5), \mathit{on}(c_6), \mathit{off}(c_7), \mathit{off}(c_8), \mathit{off}(c_(), \mathit{off}(c_{10}), \mathit{off}(c_{11})} \\
S_4 & = & \set{\mathit{off}(c_1), \mathit{off}(c_2), \mathit{on}(c_3), \mathit{on}(c_4), \mathit{off}(c_5), \mathit{on}(c_6), \mathit{off}(c_7), \mathit{off}(c_8), \mathit{off}(c_(), \mathit{off}(c_{10}), \mathit{off}(c_{11})} \\
S_5 & = & \set{\mathit{off}(c_1), \mathit{on}(c_2), \mathit{on}(c_3), \mathit{on}(c_4), \mathit{on}(c_5), \mathit{on}(c_6), \mathit{off}(c_7), \mathit{off}(c_8), \mathit{off}(c_(), \mathit{off}(c_{10}), \mathit{off}(c_{11})} \\
... &&
\end{eqnarray*}
Note that we do \emph{not} provide the spatial relation between cells. The system does not know that e.g.~cell $c_1$ is directly to the left of cell $c_2$.

We provide a sequence of templates $(\chi_1, \chi_2, ...)$ for the ECA domain. 
Our initial template $\chi_1$ is:
\begin{eqnarray*}
\phi & = & \bigsetbegin{}
T = \set{\mathit{cell}} \\
O = \set{c_1{:}\mathit{cell}, c_2{:}\mathit{cell}, ..., c_{11}{:}\mathit{cell}} \\
P = \set{\mathit{on}(\mathit{cell}), \mathit{off}(\mathit{cell}), r(\mathit{cell}, \mathit{cell})} \\
V = \set{X{:}\mathit{cell}, Y{:}\mathit{cell}, Z{:}\mathit{cell}}
\bigsetend{} \\
N_{\rightarrow} & = & 0 \\
N_{\fork} & = & 2 \\
N_B & = & 4
\end{eqnarray*}
The signature includes a binary relation $r$ on cells. This could be used as a spatial relation between neighbouring cells.
But we do not, of course, insist on this particular interpretation -- the system is free to interpret the $r$ relation any way it chooses.
The other templates $\chi_2, \chi_3, ...$ are generated by increasing the number of static rules, causal rules, and body atoms in $\chi_1$.

We applied our interpretation learning method to all $2^{2^3} = 256$ ECA rule-sets.
For Rule 110 (see Figure \ref{fig:trajectory110} above), it found the following interpretation $(\phi, I, R, C)$, where:
\begin{eqnarray*}
I &=& \bigsetbegin{}
\begin{tabular}{llll}
$\mathit{off}(c_1)$ &
$\mathit{off}(c_2)$ &
$\mathit{off}(c_3)$ &
$\mathit{off}(c_4)$ \\
$\mathit{off}(c_5)$ &
$\mathit{on}(c_6)$ &
$\mathit{off}(c_7)$ &
$\mathit{off}(c_8)$ \\
$\mathit{off}(c_9)$ &
$\mathit{off}(c_{10})$ &
$\mathit{off}(c_{11})$ &
$r(c_1, c_{11})$ \\
$r(c_2, c_1)$ &
$r(c_3, c_2)$ &
$r(c_4, c_3)$ &
$r(c_5, c_4)$ \\
$r(c_6, c_5)$ &
$r(c_7, c_6)$ &
$r(c_8, c_7)$ &
$r(c_9, c_8)$ \\
$r(c_{10}, c_9)$ &
$r(c_{11}, c_{10})$
\end{tabular}
\bigsetend{} \\
R &=& \bigsetbegin{}
r(X, Y) \wedge \mathit{on}(X) \wedge \mathit{off}(Y) \fork \mathit{on}(Y) \\
r(X, Y) \wedge r(Y, Z) \wedge \mathit{on}(X) \wedge \mathit{on}(Z) \wedge \mathit{on}(Y) \fork \mathit{off}(Y)
\bigsetend{} \\
C &=& \bigsetbegin{}
\forall X{:}\mathit{cell}, \; \mathit{on}(X) \oplus \mathit{off}(X) \\
\forall X{:}\mathit{cell}, \; \exists ! Y \; \mathit{r}(X, Y)
\bigsetend{}
\end{eqnarray*}  
The two update rules are a very compact representation of the 8 ECA updates in Figure \ref{fig:update110}:
the first rule states that if the right neighbour is on, then the target cell switches from off to on, while
the second rule states that if all three cells are on, then the target cell switches from on to off.
Here, the system uses $r(X, Y)$ to mean that cell $Y$ is immediately to the right of cell $X$.
Note that \emph{the system has constructed the spatial relation itself}. 
It was not given the spatial relation $r$ between cells.
All it was given was the sensor readings of the 11 cells.
It constructed the spatial relationship $r$ between the cells in order to make sense of the data.

\paragraph{Results}
Given the 256 ECA rules, all with the same initial configuration, we treated the trajectories as a prediction task and applied our system to it.
Our system was able to predict 249/256 correctly.
In each of the 7/256 failure cases, the \sys{} found a unified interpretation, but this interpretation produced a prediction which was not the same as the oracle. 

The complexities of the interpretations are shown in Table \ref{table:eca-complexity}.
Here, for a sample of ECA rules, we show the number of static rules, the number of causal rules, the total number of atoms in the body of all rules, the number of initial atoms, the total number of clauses (total number of static rules, causal rules, and initial atoms), and the total complexity of the interpretation. It is this final value that is minimized during search.
Note that the number of initial atoms is always 22 for all ECA tasks.
This is because there are 11 cells and each cell needs an initial $\mathit{on}$ or $\mathit{off}$, and also each cell $X$ needs a right-neighbour cell (a $Y$ such that $r(X,Y)$ to satisfy the $\exists!$ constraint on the $r$ relation).
So we require 11 + 11 = 22 initial atoms.

\begin{table}
\centering
\begin{tabular}{|l|p{1.2cm}|p{1.2cm}|p{1.2cm}|r|r|r|}
\hline
{\bf ECA rule} & {\bf \# static rules} & {\bf \# cause rules} & {\bf \# body atoms} & {\bf \# inits} & {\bf \# clauses} & {\bf complexity} \\
\hline
Rule \#0 & 0 & 1 & 0 & 22 & 23 & 24 \\
\hline
Rule \# 143 & 0 & 3 & 8 & 22 & 26 & 36 \\
\hline
Rule \#11 & 0 & 4 & 9 & 22 & 26 & 39 \\ 
\hline
Rule \#110 & 0 & 4 & 10 & 22 & 26 & 40 \\ 
\hline
Rule \# 167 & 0 & 4 & 13 & 22 & 26 & 43 \\ 
\hline 
Rule \# 150 & 0 & 4 & 16 & 22 & 26 & 46 \\ 
\hline
\end{tabular}
\caption{The complexity of the interpretations found for ECA prediction tasks}
\label{table:eca-complexity}
\end{table}

\subsection{Drum rhythms and nursery tunes}
\label{sec:rhythms-and-tunes}

We also tested our system on simple auditory perception tasks.
Here, each sensor is an auditory receptor that is tuned to listen for a particular note or drum beat.
In the tune tasks, there is one sensor for $C$, one for $D$, one for $E$, all the way to $\mathit{HighC}$. (There are no flats or sharps).
In the rhythm tasks, there is one sensor listening out for bass drum, one for snare drum, and one for hi-hat.
Each sensor can distinguish four loudness levels, between 0 and 3. 
When a note is pressed, it starts at max loudness (3), and then decays down to 0.
Multiple notes can be pressed simultaneously. 

\begin{figure}
\centering
\includegraphics[scale=0.22]{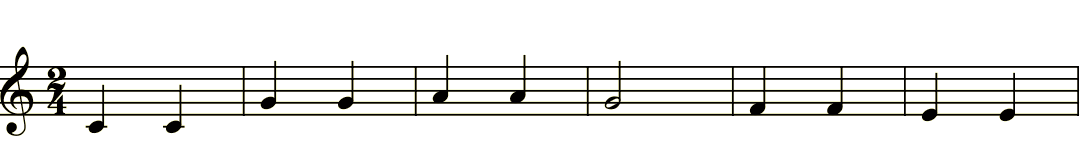}
\caption{Twinkle Twinkle Little Star tune}
\label{fig:twinkle}
\end{figure}

\begin{figure}
\centering
\includegraphics[scale=0.15]{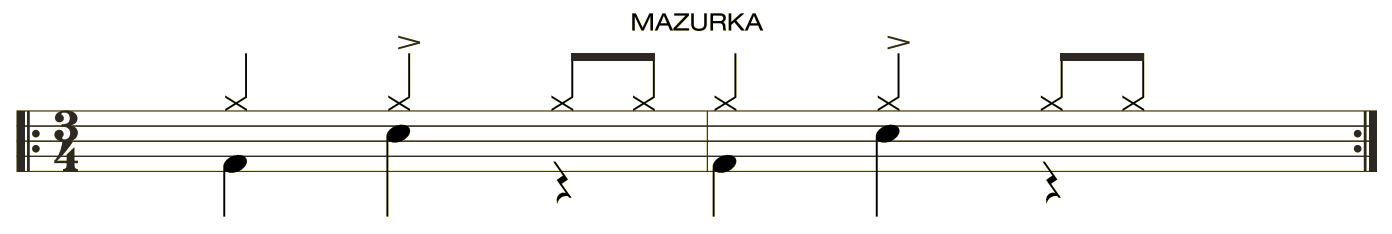}
\caption{Mazurka rhythm}
\label{fig:twinkle}
\end{figure}

For example, the \emph{Twinkle Twinkle} tune generates the following sensor readings (assuming 8 time-steps for a bar):
\begin{eqnarray*}
S_1 & = & \set{ v(s_c, 3), v(s_d, 0), v(s_e, 0), v(s_f, 0),  v(s_g, 0), v(s_a, 0), v(s_b, 0), v(s_{c*}, 0) } \\
S_2 & = & \set{ v(s_c, 2), v(s_d, 0), v(s_e, 0), v(s_f, 0),  v(s_g, 0), v(s_a, 0), v(s_b, 0), v(s_{c*}, 0) } \\
S_3 & = & \set{ v(s_c, 3), v(s_d, 0), v(s_e, 0), v(s_f, 0),  v(s_g, 0), v(s_a, 0), v(s_b, 0), v(s_{c*}, 0) } \\
S_4 & = & \set{ v(s_c, 2), v(s_d, 0), v(s_e, 0), v(s_f, 0),  v(s_g, 0), v(s_a, 0), v(s_b, 0), v(s_{c*}, 0) } \\
S_5 & = & \set{ v(s_c, 1), v(s_d, 0), v(s_e, 0), v(s_f, 0),  v(s_g, 3), v(s_a, 0), v(s_b, 0), v(s_{c*}, 0) } \\
S_6 & = & \set{ v(s_c, 0), v(s_d, 0), v(s_e, 0), v(s_f, 0),  v(s_g, 2), v(s_a, 0), v(s_b, 0), v(s_{c*}, 0) } \\
...
\end{eqnarray*}

We provided the following initial template $\chi_1$:
\begin{eqnarray*}
\phi & = & \bigsetbegin{}
T = \set{\mathit{sensor}, \mathit{finger}, \mathit{loudness}} \\
O = \set{f{:}\mathit{finger}, s_c{:}\mathit{sensor}, s_d{:}\mathit{sensor}, s_e{:}\mathit{sensor}, ..., 0{:}\mathit{loudness}, 1{:}\mathit{loudness}, ...} \\
P = \set{\mathit{part}(\mathit{finger}, \mathit{sensor}), v(\mathit{sensor}, \mathit{loudness}), r(\mathit{sensor}, \mathit{sensor}), \\
\mathit{succ}(\mathit{loudness}, \mathit{loudness}), \mathit{max}(\mathit{loudness}), \mathit{min}(\mathit{loudness})} \\
V = \set{F{:}\mathit{finger}, L{:}\mathit{loudness}, S{:}\mathit{sensor}}
\bigsetend{} \\
N_{\rightarrow} & = & 2\\
N_{\fork} & = & 3 \\
N_B & = & 2
\end{eqnarray*}
To generate the rest of the template sequence $(\chi_1, \chi_2, \chi_3, ...)$, we added additional unary predicates $p_i(\mathit{sensor})$, $q_i(\mathit{finger})$ and relational predicates $r_i(\mathit{sensor}, \mathit{sensor})$, as well as additional variables $L_i{:}\mathit{loudness}$ and $S_i{:}\mathit{sensor}$.
We also provided domain-specific knowledge of the $\mathit{succ}$ relation on loudness levels (e.g. $\mathit{succ}(0, 1), \mathit{succ}(1, 2), ...$), and we provide the spatial relation $r$ on notes: $r(s_c, s_d), r(s_d, s_e), ..., r(s_b, s_{c*})$. This is the only domain-specific knowledge given.

We tested our system on some simple rhythms (\emph{Pop Rock}, \emph{Samba}, etc.) and tunes (\emph{Twinkle Twinkle}, \emph{Three Blind Mice}, etc).
On the first two bars of \emph{Twinkle Twinkle}, it finds an interpretation with 6 rules and 26 initial atoms.
One of the rules states that when sensor $S$ satisfies predicate $p_1$, then the value of the sensor $S$ is set to the $\mathit{max}$ loudness level:
\[
p_1(S) \wedge \mathit{max}(L) \wedge v(S, L_2) \fork v(S, L) 
\]
This rule states that when sensor $S$ satisfies $p_2$, then the value decays:
\[
p_2(S) \wedge \mathit{succ}(L, L2) \wedge v(S, L_2) \fork v(S, L) 
\]
Clearly, $p_1$ and $p_2$ are exclusive unary predicates used to determine whether a note is currently being pressed or not.

The next rule states that when the finger $F$ satisfies predicate $q_1$, then the note which the finger is on is pressed:
\[
q_1(F) \wedge \mathit{part}(F, S) \wedge p_2(S) \fork p_1(S)
\]
Here, the system is using $q_1$ to indicate whether or not the finger is down. It uses the other predicates $q_2, q_3, ...$ to indicate which state the finger is in (and hence which note the finger should be on), and the other rules to indicate when to transition from one state to another.

\paragraph{Results}
Recall that our accuracy metric is stringent and only counts a prediction as accurate if \emph{every} sensor's value is predicted correctly.
In the rhythm and music domain, this means the \sys{} must correctly predict the loudness value (between 0 and 3) for each of the sound sensors. There are 8 sensors for tunes and 3 sensors for rhythms. 

When we tested the \sys{} on the 20 drum rhythms and 10 nursery tunes, our system was able to predict 22/30 correctly.
The complexities of the interpretations are shown in Table \ref{table:rhythm-complexity}.
Note that the interpretations found are large and complex programs by the standards of state of the art ILP systems.
In \emph{Mazurka}, for example, the interpretation contained 16 update rules with 44 body atoms.
In \emph{Three Blind Mice}, the interpretation contained 10 update rules and 34 initialisation atoms making a total of 44 clauses.

In the 8 cases where the \sys{} failed to predict correctly, this was because the system failed to find a unified interpretation of the sensory sequence. It was not that the system found an interpretation which produced the wrong prediction.
Rather, in the 8 failure cases, it was simply unable to find a unified interpretation within the memory and time limits.
In the ECA tasks, by contrast, the system always found some unified interpretation for each of the 256 tasks, but some of these interpretations produced the wrong prediction.

\begin{table}
\centering
\begin{tabular}{|l|p{1.2cm}|p{1.2cm}|p{1.2cm}|r|r|r|}
\hline
{\bf Task} & {\bf \# static rules} & {\bf \# cause rules} & {\bf \# body atoms} & {\bf \# inits} & {\bf \# clauses} & {\bf complexity} \\
\hline
Twinkle Twinkle & 2 & 4 & 9 & 26 & 32 & 45 \\
\hline
Eighth Note Drum Beat & 4 & 8 & 29 & 13 & 25 & 62  \\
\hline
Stairway to Heaven & 4 & 8 & 30 & 13 & 25 & 63 \\ 
\hline
Three Blind Mice & 2 & 8 & 17 & 34 & 44 & 69 \\ 
\hline
Twist & 4 & 12 & 40 & 16 & 32 & 84 \\ 
\hline
Mazurka & 4 & 12 & 44 & 14 & 30 & 86 \\ 
\hline
\end{tabular}
\caption{The complexity of the interpretations found for rhythm and tune prediction tasks}
\label{table:rhythm-complexity}
\end{table}

\subsection{\emph{Seek Whence} and C-test sequence induction intelligence tests}
\label{sec:seek-whence}

Hofstadter introduced the \define{Seek Whence}\footnote{The name is a pun on ``sequence''. See also the related Copycat domain \cite{mitchell1993analogy}.} domain in \cite{hofstadter2008fluid}. 
The task is, given a sequence $s_1, ..., s_t$ of symbols, to predict the next symbol $s_{t+1}$.
Typical \emph{Seek Whence} tasks include\footnote{Hofstadter used natural numbers, but we transpose the sequences to letters, to bring them in line with the Thurstone letter completion problems \cite{thurstone1941factorial} and the C-test \cite{hernandez1998formal}.}:
\begin{itemize}
\item
\sequence{b, b, b, c, c, b, b, b, c, c, b, b, b, c, c}
\item
\sequence{a, f, b, f, f, c, f, f, f, d, f, f}
\item
\sequence{b, a, b, b, b, b, b, c, b, b, d, b, b, e, b}
\end{itemize}
Hofstadter called the third sequence the ``theme song'' of the \emph{Seek Whence} project because of its difficulty.
There is a ``perceptual mirage'' in the sequence because of the sub-sequence of five $b$'s in a row that makes it hard to see the intended pattern: $(b,x,b)^*$ for ascending $x$.

It is important to note that these tasks, unlike the tasks in the ECA domain or in the rhythm and music domain, have a certain subjective element.
There are always many different ways of interpreting a finite sequence.
Given that these different interpretations will provide different continuations, why privilege some continuations over others?

When Hern\'andez-Orallo introduced the C-test \cite{hernandez1998formal,hernandez2000beyond,hernandez2016computer,hernandez2017measure}, one of his central motivations was to address this ``subjectivity'' objection via the concept of \emph{unquestionability}.
If we are given a particular programming language for generating sequences, then a sequence $s_{1:T}$ is \define{unquestionable} if it is not the case that the smallest program $\pi$ that generates $s_{1:T}$ is rivalled by another program $\pi'$ that is almost as small, where $\pi$ and $\pi'$ have different continuations after $T$ time-steps.

Consider, for example, the sequence \sequence{a, b, b, c, c}
This sequence is highly questionable because there are two interpretations which are very similar in length (according to most programming languages), one of which parses the sequence as $(a), (b, b), (c, c, c), (d, d, d, d), ...$ and the other of which parses the sequence as a sequence of pairs $(a, b), (b, c), (c, d), ...$.
Hern\'andez-Orallo generated the C-test sequences by enumerating programs from a particular domain-specific language, executing them to generate a sequence, and then restricting the computer-generated sequences to those that are unquestionable.

For our set of sequence induction tasks, we combined sequences from Hofstadter's \emph{Seek Whence} dataset (transposed from numbers to letters) together with sequences from the C-test.
The C-test sequences are unquestionable by construction, and we also observed (by examining the size of the smallest interpretations) that Hofstadter's sequences were unquestionable with respect to \logic{}. 
This goes some way to answering the ``subjectivity'' objection\footnote{Some may still be concerned that the definition of unquestionability is relative to a particular domain-specific language, and the Kolmogorov complexity of a sequence depends on the choice of language. Hern\'andez-Orallo \cite{hernandez2017measure} discusses this issue at length.}.

There have been a number of attempts to implement sequence induction systems using domain-specific knowledge of the types of sequence to be encountered.
Simon et al \cite{simon1963human} implemented the first program for solving Thurstone's letter sequences \cite{thurstone1941factorial}.
%
Meredith \cite{meredith1986seek} and Hofstadter \cite{hofstadter2008fluid} also used domain-specific knowledge: 
after observing various types of commonly recurring patterns in the \emph{Seek Whence} sequences, they hand-crafted a set of procedures to detect the patterns. 
Although their search algorithm is general, the patterns over which it is searching are hand-coded and domain-specific.

If solutions to sequence induction or IQ tasks are to be useful in general models of cognition, it is essential that we do not provide domain-specific solutions to those tasks. 
As Hern{\'a}ndez-Orallo et al \cite{hernandez2016computer} argue, `` In fact, for most approaches the system does not learn to solve the problems but it is programmed to solve the problems. In other words, the task is hard-coded into the program and it can be easier to become `superhuman' in many specific tasks, as happens with chess, draughts, some kinds of planning, and many other tasks. But humans are not programmed to do intelligence tests.''
What we want is a \emph{general-purpose domain-agnostic} perceptual system that can solve sequence induction tasks ``out of the box'' \emph{without hard-coded domain-specific knowledge} \cite{besold2015artificial}.

The \sys{} described in this chapter is just such a general-purpose domain-agnostic perceptual system. 
We tested it on 30 sequences (see Figure \ref{diag:sw}), and it got $76.6\%$ correct (23 out of 30 correct, 3 out of 30 incorrect and 4 out of 30 timeout).  

\begin{figure}
\centering
\begin{tabular}{|l|l|}
\hline
\sequence{a,a,b,a,b,c,a,b,c,d,a} & 
\sequence{a,b,c,d,e}\\
\sequence{b,a,b,b,b,b,b,c,b,b,d,b,b,e} &
\sequence{a,b,b,c,c,c,d,d,d,d,e}\\
\sequence{a,f,e,f,a,f,e,f,a,f,e,f,a} &
\sequence{b,a,b,b,b,c,b,d,b,e}\\
\sequence{a,b,b,c,c,d,d,e,e} &
\sequence{a,b,c,c,d,d,e,e,e,f,f,f}\\
\sequence{f,a,f,b,f,c,f,d,f} &
\sequence{a,f,e,e,f,a,a,f,e,e,f,a,a}\\
\sequence{b,b,b,c,c,b,b,b,c,c,b,b,b,c,c} &
\sequence{b,a,a,b,b,b,a,a,a,a,b,b,b,b,b}\\
\sequence{b,c,a,c,a,c,b,d,b,d,b,c,a,c,a} &
\sequence{a,b,b,c,c,d,d,e,e,f,f}\\
\sequence{a,a,b,a,b,c,a,b,c,d,a,b,c,d,e} &
\sequence{b,a,c,a,b,d,a,b,c,e,a,b,c,d,f}\\
\sequence{a,b,a,c,b,a,d,c,b,a,e,d,c,b} &
\sequence{c,b,a,b,c,b,a,b,c,b,a,b,c,b}\\
\sequence{a,a,a,b,b,c,e,f,f} &
\sequence{a,a,b,a,a,b,c,b,a,a,b,c,d,c,b}\\
\sequence{a,a,b,c,a,b,b,c,a,b,c,c,a,a,a} &
\sequence{a,b,a,b,a,b,a,b,a}\\
\sequence{a,c,b,d,c,e,d} &
\sequence{a,c,f,b,e,a,d}\\
\sequence{a,a,f,f,e,e,d,d} &
\sequence{a,a,a,b,b,b,c,c}\\
\sequence{a,a,b,b,f,a,b,b,e,a,b,b,d} &
\sequence{f,a,d,a,b,a,f,a,d,a,b,a}\\
\sequence{a,b,a,f,a,a,e,f,a} &
\sequence{b,a,f,b,a,e,b,a,d}\\
\hline
\end{tabular}
\caption{Sequences from \emph{Seek Whence} and the C-test}
\label{diag:sw}
\end{figure}

For the letter sequence induction problems, we provide the initial template $\chi_1$:
\begin{eqnarray*}
\phi & = & \bigsetbegin{}
T = \set{\mathit{sensor}, \mathit{cell}, \mathit{letter}} \\
O = \set{s{:}\mathit{sensor}, c_1{:}\mathit{cell}, l_{a}{:}\mathit{letter}, l_{b}{:}\mathit{letter}, l_{c}{:}\mathit{letter}, ...} \\
P = \set{\mathit{value}(\mathit{sensor}, \mathit{letter}), \mathit{part}(\mathit{sensor}, \mathit{cell}), p(\mathit{cell}, \mathit{letter}), q_1(\mathit{cell}), r(\mathit{cell}, \mathit{cell})} \\
V = \set{X{:}\mathit{sensor}, Y{:}\mathit{cell}, Y_2{:}\mathit{cell}, L{:}\mathit{letter}, L_2{:}\mathit{letter}}
\bigsetend{} \\
N_{\rightarrow} & = & 1 \\
N_{\fork} & = & 2 \\
N_B & = & 3
\end{eqnarray*}
As we iterate through the templates $(\chi_1, \chi_2, \chi_3, ...)$, we increase the number of objects, the number of fluent and permanent predicates, the number of static rules and causal rules, and the number of atoms allowed in the body of a rule.

The one piece of domain-specific knowledge we inject is the successor relation between the letters $l_a$, $l_b$, $l_c$, ...
We provide the $\mathit{succ}$ relation with $\mathit{succ}(l_a, l_b), \mathit{succ}(l_b, l_c), ...$
Please note that this knowledge does not \emph{have} to be given to the system. 
We verified it is possible for the system to learn the successor relation on a simpler task and then \emph{reuse} this information in subsequent tasks. 
We plan to do more continual learning from curricula in future work.

\begin{figure}
\centering
\includegraphics[scale=0.75]{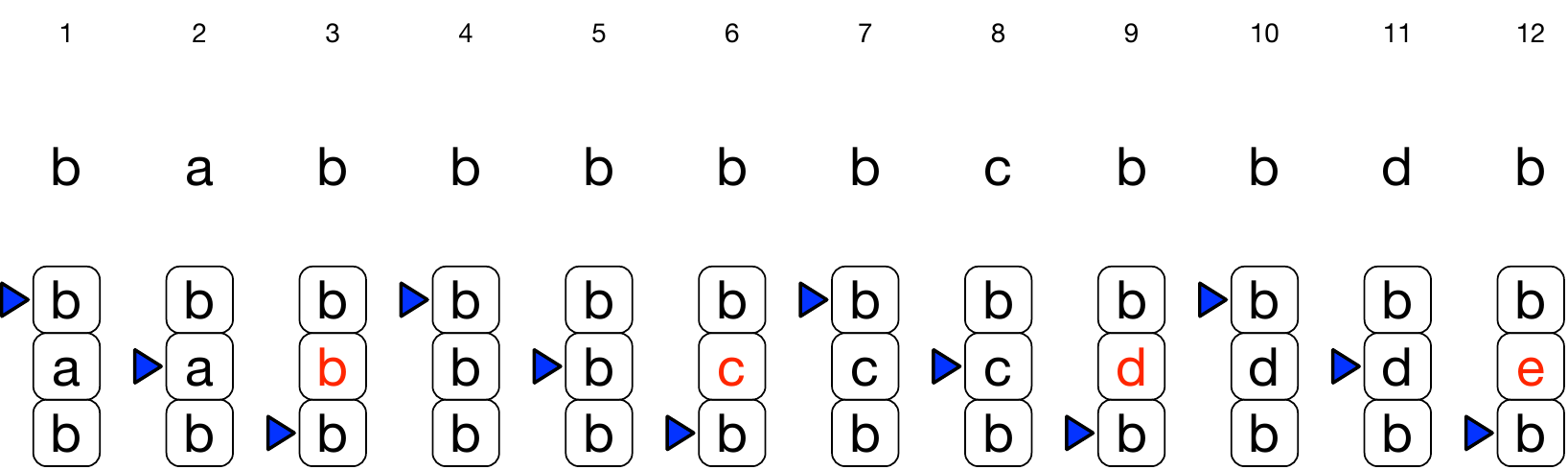}
\caption[Our interpretation of the ``theme song'' \emph{Seek Whence} sequence]{A visualization of the \sys{}'s interpretation of the ``theme song'' \emph{Seek Whence} sequence \sequence{b, a, b, b, b, b, b, c, b, b, d, b, b, e, b, b}. We show the trace $\tau(\theta) = A_1, A_2, ..., $ of this interpretation for the first 12 time steps.
The $t$'th column represents the state at time $t$.
Each column contains the time index $t$, the sensor reading $S_t$, the values of the three latent objects $c_1, c_2, c_3$ at time $t$, and the position of the sensor $s$ at $t$.
The only moving object is the sensor, represented by a  triangle, that moves between the three latent objects from top to bottom and then repeats. 
Note that the middle object $c_2$'s value changes when the sensor passes over it; we change the color of the object's letter to indicate when the object's value has changed.}
\label{fig:theme-song}
\end{figure}

We illustrate our system on the ``theme song'' of the \emph{Seek Whence} project: \sequence{b, a, b, b, b, b, b, c, b, b, d, b, b, e, b, b}.
Let the sensory sequence be $S_{1:16}$ where:
\[
\begin{tabular}{lll}
$S_1 = \set{ \mathit{value}(s, l_b)}$ & $S_2 = \set{ \mathit{value}(s, l_a)}$ & $S_3 = \set{ \mathit{value}(s, l_b)}$ \\
$S_4 = \set{ \mathit{value}(s, l_b)}$ & $S_5 = \set{ \mathit{value}(s, l_b)}$ & $S_6 = \set{ \mathit{value}(s, l_b)}$ \\
$S_7 = \set{ \mathit{value}(s, l_b)} $ & $S_8 = \set{ \mathit{value}(s, l_c)}$ & $S_9 = \set{ \mathit{value}(s, l_b)}$ \\
$S_{10}= \set{ \mathit{value}(s, l_b)}$ & $S_{11} = \set{ \mathit{value}(s, l_d)}$ & $S_{12} = \set{ \mathit{value}(s, l_b)}$ \\
$S_{13}= \set{ \mathit{value}(s, l_b)}$ & $S_{14} = \set{ \mathit{value}(s, l_e)}$ & $S_{15} = \set{ \mathit{value}(s, l_b)}$ \\
$S_{16}= \set{ \mathit{value}(s, l_b)}$ & ... & ...
\end{tabular}
\]
When our system is run on this sensory input, the first few templates are unable to find a solution. 
The first template that is expressive enough to admit a solution is one where there are three latent objects $c_1, c_2, c_3$.
The first interpretation found is $(\phi, I, R, C)$ where:
\begin{eqnarray*}
I & = & \bigsetbegin{}
\begin{tabular}{llll}
$p(c_1, l_b)$ &  $p(c_2, l_b)$ & $p(c_3, l_a)$ & $q_1(c_3)$ \\
$q_2(c_1)$ & $q_2(c_2)$ & $r(c_1, c_3)$ & $r(c_3, c_2)$ \\
$r(c_2, c_1)$ & $\mathit{part}(s, c_1)$ & & \\
\end{tabular}
\bigsetend{} \\
R & = & \bigsetbegin{}
\mathit{part}(X, Y) \wedge p(Y, L) \rightarrow \mathit{value}(X, L) \\
r(Y, Y_2) \wedge \mathit{part}(X, Y) \fork \mathit{part}(X, Y_2) \\
\mathit{part}(X, Y) \wedge q_1(Y) \wedge \mathit{succ}(L, L_2) \wedge p(Y, L) \fork p(Y, L_2)
\bigsetend{} \\
C & = & \bigsetbegin{}
\forall X{:}\mathit{sensor}, \; \exists ! L \; \mathit{value}(X, L) \\
\forall Y{:}\mathit{cell}, \; \exists ! L \; \mathit{p}(Y, L) \\
\forall Y{:}\mathit{cell} \; q_1(Y) \oplus q_2(Y) \\
\forall Y{:}\mathit{cell}, \; \exists ! Y_2 \; r(Y, Y_2)
\bigsetend{}
\end{eqnarray*}
In this interpretation, the sensor moves between the three latent objects in the order $c_1, c_3, c_2, c_1, c_3, c_2, ...$
The two unary predicates $q_1$ and $q_2$ are used to divide the latent objects into two types.
Effectively, $q_1$ is interpreted as an ``object that increases its letter'' while $q_2$ is interpreted as a ``static object''.
The static rule states that the sensor inherits its value from the $p$-value of the object it is on.
The causal rule $r(Y, Y_2) \wedge \mathit{part}(X, Y) \fork \mathit{part}(X, Y_2)$ states that the sensor moves from left to right along the latent objects. 
The causal rule $\mathit{part}(X, Y) \wedge q_1(Y) \wedge \mathit{succ}(L, L_2) \wedge p(Y, L) \fork p(Y, L_2)$ states that $q_1$ objects increase their $p$-value when a sensor is on them.
This is an intelligible and satisfying interpretation of the sensory sequence.
See Diagram \ref{fig:theme-song}.

\paragraph{Results}
Given the 30 \emph{Seek Whence} sequences, we treated the trajectories as a prediction task and applied our system to it.
Our system was able to predict 23/30 correctly.
For the 7 failure cases, 4 of them were due to the system not being able to find any unified interpretation within the memory and time limits, while in 3 of them, the system found a unified interpretation that produced the ``incorrect'' prediction.
The complexities of the interpretations are shown in Table \ref{table:sw-complexity}.
\begin{table}
\centering
\begin{tabular}{|l|p{1.2cm}|p{1.2cm}|p{1.2cm}|r|r|r|}
\hline
{\bf Sequence} & {\bf \# static rules} & {\bf \# cause rules} & {\bf \# body atoms} & {\bf \# inits} & {\bf \# clauses} & {\bf complexity} \\
\hline
abcde... & 0 & 1 & 1 & 7 & 8 & 10 \\ 
\hline
fafbfcfdf... & 1 & 2 & 6 & 7 & 10 & 18 \\ 
\hline
babbbbbcbbdbbe... & 1 & 2 & 6 & 14 & 17 & 25 \\
\hline
aababcabcdabcde... & 3 & 5 & 19 & 7 & 15 & 39 \\ 
\hline
abccddeeefff... & 3 & 5 & 21 & 8 & 16 & 42 \\ 
\hline
baabbbaaaabbbbb... & 3 & 5 & 23 & 7 & 15 & 43 \\ 
\hline
\end{tabular}
\caption{The complexity of the interpretations found for \emph{Seek Whence} prediction tasks}
\label{table:sw-complexity}
\end{table}
The first key point we want to emphasise here is that our system was able to achieve human-level performance\footnote{See Meredith \cite{meredith1986seek} for empirical results 25 students  on the``Blackburn dozen'' \emph{Seek Whence} problems.}  on these tasks without hand-coded domain-specific knowledge.
This is a \emph{general} system for making sense of sensory data that, when applied to the \emph{Seek Whence} domain\footnote{The only domain-specific information provided is the $\mathit{succ}$ relation on letters.}, is able to solve these particular problems.
The second point we want to stress is that our system did not learn to solve these sequence induction tasks after seeing many previous examples\footnote{Machine learning approaches to these tasks need thousands of examples before they can learn to predict. See for example \cite{barrett2018measuring}.}. 
On the contrary: our system had never seen any such sequences before; it confronts each sequence \emph{de novo}, without prior experience.
This system is, to the best of our knowledge, the first such general system that is able to achieve such a result.

\subsection{Binding tasks}
\label{sec:probe-task-binding}

We wanted to see whether our system could handle traditional problems from cognitive science ``out of the box'', without needing additional task-specific information. We used probe tasks to evaluate two key issues: binding and occlusion.

The binding problem \cite{holcombe2009binding} is the task of recognising that information from different sensory modalities should be collected together as different aspects of a single external object.
For example, you hear a buzzing and a siren in your auditory field and you see an insect and an ambulance in your visual field.
How do you associate the buzzing and the insect-appearance as aspects of one object, and the siren and the ambulance appearance as aspects of a separate object?

To investigate how our system handles such binding problems,  we tested it on the following multi-modal variant of the ECA described above.
Here, there are two types of sensor. The light sensors have just two states: black and white, while the touch sensors have four states: fully un-pressed (0), fully pressed (3), and two intermediate states (1, 2).
After a touch sensor is fully pressed (3), it slowly depresses, going from states 2 to 1 to 0 over 3 time-steps.
In this example, we chose Rule 110 (the Turing-complete ECA rule) with the same initial configuration as in Figure \ref{fig:trajectory110}, as described earlier.
In this multi-modal variant, there are 11 light sensors, one for each cell in the ECA, and two touch sensors on cells 3 and 11.
See Figure \ref{fig:binding}.

Suppose we insist that the type signature contains no binary relations connecting any of the sensors together.
Suppose there is no relation in the given type signature between light sensors, no relation between touch sensors, and no relation between light sensors and touch sensors.
Now, in order to satisfy the constraint of spatial unity, there must be some indirect connection between any two sensors.
But if there are no direct relations between the sensors, \emph{the only way our system can satisfy the constraint of spatial unity is by positing latent objects, directly connected to each other, that the sensors are connected to}.
Thus the latent objects are the intermediaries through which the various sensors are indirectly connected.

For the binding tasks, we started with the initial template $\chi_1$:
\begin{eqnarray*}
\phi & = & \bigsetbegin{}
T = \set{\mathit{cell}, \mathit{light}, \mathit{touch}, \mathit{int}} \\
O = \set{c_1{:}\mathit{cell}, c_2{:}\mathit{cell}, ..., c_{11}{:}\mathit{cell}, l_1{:}\mathit{light}, l_2{:}\mathit{light}, ..., l_{11}{:}\mathit{light}, t_1{:}\mathit{touch}, \\
t_2{:}\mathit{touch}, 0{:}\mathit{int}, 1{:}\mathit{int}, ...} \\
P = \set{\mathit{black}(\mathit{light}), \mathit{white}(\mathit{light}), \mathit{value}(\mathit{touch}, \mathit{int}), \mathit{on}(\mathit{cell}), \\
\mathit{off}(\mathit{cell}), r(\mathit{cell}, \mathit{cell}), \mathit{in}_L(\mathit{light}, \mathit{cell}), \mathit{in}_T(\mathit{touch}, \mathit{cell})} \\
V = \set{C{:}\mathit{cell}, X{:}\mathit{touch}, Y{:}\mathit{light}, L{:}\mathit{int}}
\bigsetend{} \\
N_{\rightarrow} & = & 4 \\
N_{\fork} & = & 4 \\
N_B & = & 4
\end{eqnarray*}
As we iterate through the templates $(\chi_1, \chi_2, \chi_3, ...)$, we increase the number of predicates, the number of variables, the number of static rules and causal rules, and the number of atoms allowed in the body of a rule.

Given the template sequence $(\chi_1, \chi_2, \chi_3, ...)$, our system found the following interpretation $(\phi, I, R, C)$, where:
\begin{eqnarray*}
I &=& \bigsetbegin{}
\begin{tabular}{llll}
$\mathit{off}(c_1)$ &
$\mathit{off}(c_2)$ &
$\mathit{off}(c_3)$ &
$\mathit{off}(c_4)$ \\
$\mathit{off}(c_5)$ &
$\mathit{on}(c_6)$ &
$\mathit{off}(c_7)$ &
$\mathit{off}(c_8)$ \\
$\mathit{off}(c_9)$ &
$\mathit{off}(c_{10})$ &
$\mathit{off}(c_{11})$ &
$r(c_1, c_{11})$ \\
$r(c_2, c_1)$ &
$r(c_3, c_2)$ &
$r(c_4, c_3)$ &
$r(c_5, c_4)$ \\
$r(c_6, c_5)$ &
$r(c_7, c_6)$ &
$r(c_8, c_7)$ &
$r(c_9, c_8)$ \\
$r(c_{10}, c_9)$ &
$r(c_{11}, c_{10})$ &
$\mathit{in}_L(l_1, c_1)$ &
$\mathit{in}_L(l_2, c_2)$ \\
$...$ &
$\mathit{in}_L(l_{11}, c_{11})$ &
$\mathit{in}_T(t_1, c_3)$ &
$\mathit{in}_T(t_2, c_{11})$
\end{tabular}
\bigsetend{} \\
R &=& \bigsetbegin{}
r(C_1, C_2) \wedge \mathit{on}(C_2) \wedge \mathit{off}(C_1) \fork \mathit{on}(C_1) \\
r(C_1, C_2) \wedge r(C_2, C_3) \wedge \mathit{on}(C_1) \wedge \mathit{on}(C_3) \wedge \mathit{on}(C_2) \fork \mathit{off}(C_2) \\
\mathit{touch}(X, L_1) \wedge \mathit{min}(L_1) \wedge p(X, L_2) \fork p(X, L_1)  \\
\mathit{touch}(X, L_1) \wedge \mathit{succ}(L_2, L_1) \wedge p(X, L_1) \fork p(X, L_2)  \\
\mathit{in}_T(X, C) \wedge \mathit{on}(C) \wedge \mathit{max}(L) \rightarrow \mathit{value}(X, L) \\
\mathit{in}_T(X, C) \wedge \mathit{off}(C) \wedge p(X, L) \rightarrow \mathit{value}(X, L) \\
\mathit{in}_L(Y, C) \wedge \mathit{on}(C) \rightarrow \mathit{black}(Y) \\
\mathit{in}_L(Y, C) \wedge \mathit{off}(C) \rightarrow \mathit{white}(Y)
\bigsetend{} \\
C &=& \bigsetbegin{}
\forall C{:}\mathit{cell}, \; \mathit{on}(C) \oplus \mathit{off}(C) \\
\forall C{:}\mathit{cell}, \; \exists ! C_2 \; \mathit{r}(C, C_2) \\
\forall X{:}\mathit{touch}, \; \exists ! C \; \mathit{in}_T(X, C) \\
\forall X{:}\mathit{touch}, \; \exists ! L \; \mathit{value}(X, L) \\
\forall Y{:}\mathit{light}, \; \mathit{black}(Y) \oplus \mathit{white}(Y) \\
\forall Y{:}\mathit{light}, \; \exists ! C \; \mathit{in}_L(X, C)
\bigsetend{}
\end{eqnarray*}
Here, the cells $c_1, ..., c_{11}$ are directly connected via the $r$ relation, and the light and touch sensors are connected to the cells via the $in_L$ and $in_T$ relations. Thus all the sensors are indirectly connected. For example, light sensor $l_1$ is indirectly connected to touch sensor $t_1$ via the chain of relations $in_L(l_1, c_1),  r(c_1, c_2), r(c_2, c_3), in_T(t_1, c_3)$.
We stress that we did not need to write special code in order to get the system to satisfy the binding problem.
Rather, \emph{the binding problem is satisfied automatically, as a side-effect of satisfying the spatial unity condition}.

We ran 20 multi-modal binding experiments, with different ECA rules, different initial conditions, and the touch sensors attached to different cells. The results are shown in Table \ref{table:probe-results}.

\begin{figure}
\begin{center}
\begin{tikzpicture}[b/.style={draw, minimum size=3mm,   
       fill=black},w/.style={draw, minimum size=3mm},
       m/.style={matrix of nodes, column sep=1pt, row sep=1pt, draw, label=below:#1}, node distance=1pt]

\matrix (A) [m=110]{
$l_1$&$l_2$&${\color{red}l_3}$&$l_4$&$l_5$&$l_6$&$l_7$&$l_8$&$l_9$&$l_{10}$&${\color{blue}l_{11}}$&$t_1$&$t_2$\\
W&W&{\color{red}W}&W&W&B&W&W&W&W&{\color{blue}W}&{\color{red}0}&{\color{blue}0}\\
W&W&{\color{red}W}&W&B&B&W&W&W&W&{\color{blue}W}&{\color{red}0}&{\color{blue}0}\\
W&W&{\color{red}W}&B&B&B&W&W&W&W&{\color{blue}W}&{\color{red}0}&{\color{blue}0}\\
W&W&{\color{red}B}&B&W&B&W&W&W&W&{\color{blue}W}&{\color{red}3}&{\color{blue}0}\\
W&B&{\color{red}B}&B&B&B&W&W&W&W&{\color{blue}W}&{\color{red}3}&{\color{blue}0}\\
B&B&{\color{red}W}&W&W&B&W&W&W&W&{\color{blue}W}&{\color{red}2}&{\color{blue}0}\\
B&B&{\color{red}W}&W&B&B&W&W&W&W&{\color{blue}B}&{\color{red}1}&{\color{blue}3}\\
W&B&{\color{red}W}&B&B&B&W&W&W&B&{\color{blue}B}&{\color{red}0}&{\color{blue}3}\\
B&B&{\color{red}B}&B&W&B&W&W&B&B&{\color{blue}B}&{\color{red}3}&{\color{blue}3}\\
W&W&{\color{red}W}&B&B&B&W&B&B&W&{\color{blue}W}&{\color{red}2}&{\color{blue}2}\\
?&?&?&?&?&?&?&?&?&?&?&?&?\\
};
\end{tikzpicture}  
\end{center}
\caption[A multi-modal trace of ECA rule 110 with light sensors and touch sensors]{A multi-modal trace of ECA rule 110 with eleven light sensors (left) $l_1, ..., l_{11}$ and two touch sensors (right) $t_1, t_2$ attached to cells 3 and 11. Each row represents the states of the sensors for one time-step. For this prediction task, the final time-step is held out.}
\label{fig:binding}
\end{figure}

\begin{table}
\centering
\begin{tabular}{|l|r|r|r|r|}
\hline
{\bf Domain} & {\bf \# Tasks} & {\bf Memory} & {\bf Time} & {\bf\% Correct} \\
\hline
Multi-modal binding & 20 & 1003.2 & 2.4 & 85.0\% \\
\hline
Occlusion & 20 & 604.3 & 2.3 & 90.0 \% \\
\hline
\end{tabular}
\caption[The two types of probe task]{The two types of probe task. We show mean memory in megabytes and mean solution time in hours.}
\label{table:probe-results}
\end{table}

\subsection{Occlusion tasks}
\label{sec:probe-task-binding}

Neural nets that predict future sensory data conditioned on past sensory data struggle to solve occlusion tasks because it is hard to inject into them the prior knowledge that objects persist over time.
Our system, by contrast, was designed to posit latent objects that persist over time.

\begin{figure}
\centering
\includegraphics[scale=0.50]{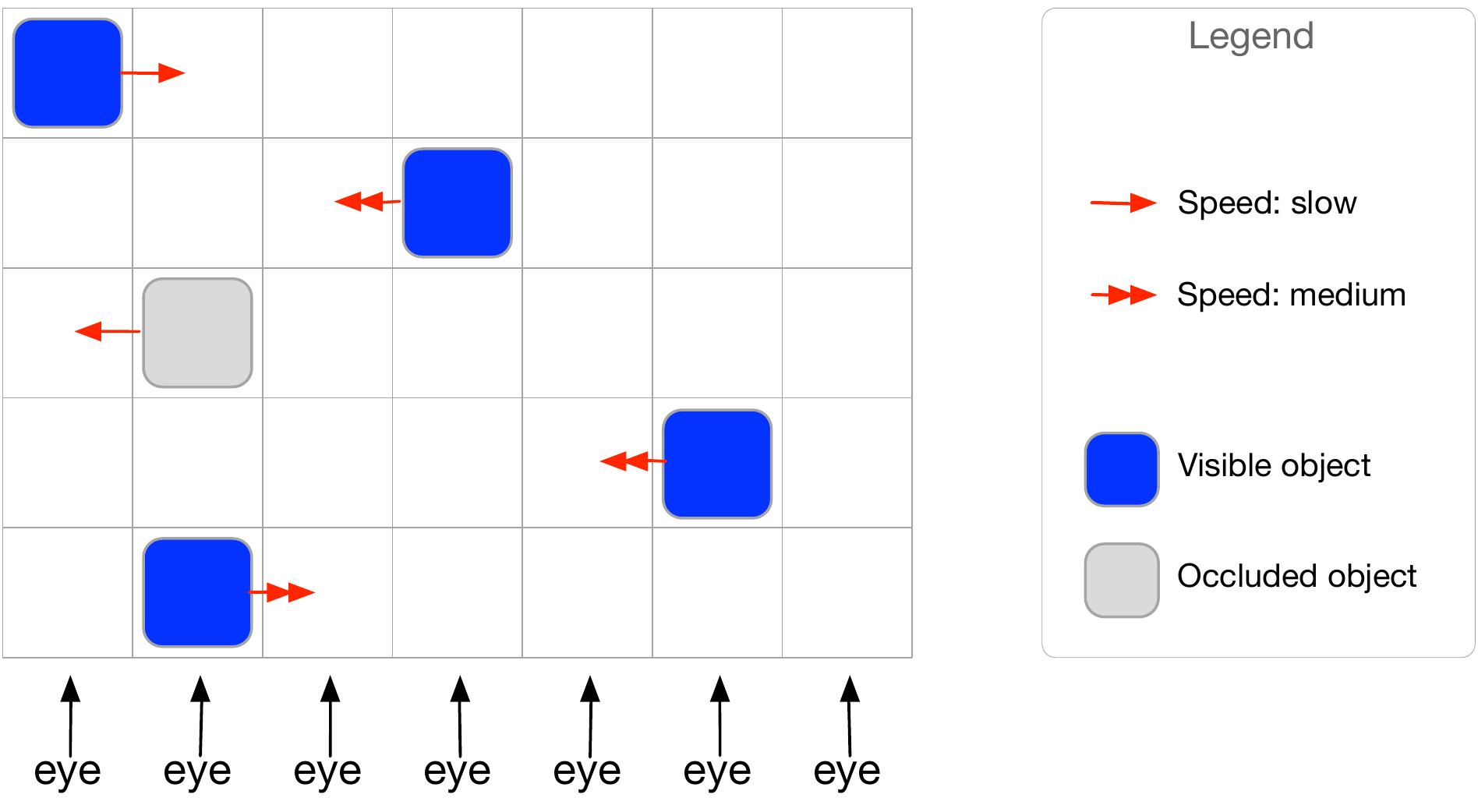}
\caption{An occlusion task}
\label{fig:occlusion}
\end{figure}

To test our system's ability to solve occlusion problems, we generated a set of tasks of the following form:
there is a 2D grid of cells in which objects move horizontally.
Some move from left to right, while others move from right to left, with wrap around when they get to the edge of a row.
The objects move at different speeds. 
Each object is placed in its own row, so there is no possibility of collision.
There is an ``eye'' placed at the bottom of each column, looking up. 
Each eye can only see the objects in the column it is placed in.
An object is occluded if there is another object below it in the same column.
See Figure \ref{fig:occlusion}.

The system receives a sensory sequence consisting of the positions of the moving objects whenever they are visible.
The positions of the objects when they are occluded is used as held-out test data to verify the predictions of the model.
This is an imputation task.

For the occlusion tasks, we provide the initial template $\chi_1$:
\begin{itemize}
\item
$T = \set{\mathit{cell}, \mathit{mover}}$
\item
$O = \set{c_{1,1}{:}\mathit{cell}, ..., c_{7,5}{:}\mathit{cell}, m_1{:}\mathit{mover}, ..., m_5{:}\mathit{mover}}$
\item
$P = \set{\mathit{in}(\mathit{mover}, \mathit{cell}), \mathit{right}(\mathit{cell}, \mathit{cell}), \mathit{below}(\mathit{cell}, \mathit{cell}), \\
p_1(\mathit{mover}), p_2(\mathit{mover}), p_3(\mathit{mover}), p_4(\mathit{mover})}
$
\end{itemize}
We provide the $\mathit{right}$ and $\mathit{below}$ predicates defining the 2D relation between grid cells.
The system is free to interpret the $p_i$ predicates any way it desires.

Here is a sample of the rules generated to solve one of the tasks:
\begin{eqnarray*}
&& p_1(X) \wedge \mathit{right}(C_1, C_2) \wedge \mathit{in}(X, C_1) \fork \mathit{in}(X, C_2) \\
&& p_2(X) \wedge p_3(X) \fork p_4(X) \\
&& p_2(X) \wedge p_4(X) \fork p_3(X) \\
&& p_2(X) \wedge  \mathit{right}(C_1, C_2) \wedge p_4(X) \wedge \mathit{in}(X, C_2) \fork \mathit{in}(X, C_1)
\end{eqnarray*}
The rules describe the behaviour of two types of moving objects.
An object of type $p_1$ moves right one cell every time-step.
An object of type $p_2$ moves left every two time-steps.
It uses the state predicates $p_3$ and $p_4$ as \emph{counters} to determine when it should move left and when it should remain where it is.

We generated 20 occlusion tasks by varying the size of the grid, the number of moving objects, their direction and speed.
Our system was able to solve these tasks without needing additional domain-specific information.
The results are shown in Table \ref{table:probe-results}.

\section{Empirical comparisons with other approaches}
\label{sec:results}

In this section, we evaluate our system experimentally and attempt to establish the following claims.
First, we claim the test domains of Section \ref{sec:experiments} represent a challenging set of tasks.
We show that these domains are challenging by providing baselines that are unable to interpret the sequences.
Second, we claim our system is general in that it can handle retrodiction and imputation as easily as it can handle prediction tasks.
We show in extensive tests that results for retrodicting earlier values and imputing intermediate values are comparable with results for predicting future values.
Third, we claim that the various features of the system (the unity conditions and the cost minimization procedure) are essential to the success of the system.
In ablation tests, where individual features are removed, the system performs significantly worse.
Fourth, we claim that the particular program synthesis technique we use is efficient when compared with state-of-the-art program synthesis methods. 
Specifically, we show how ILASP (a state-of-the-art ILP system \cite{law2014inductive,law2015learning,law2016iterative,law2018complexity}) is capable of solving some of the smaller tasks, but struggles for the larger tasks. 

\subsection{Our domains are challenging for existing baselines}
\label{sec:baselines}

To evaluate whether our domains are indeed sufficiently challenging, we compared our system against four baselines.
The first \define{constant} baseline always predicts the same constant value for every sensor for each time-step.
The second \define{inertia} baseline always predicts that the final hidden time-step equals the penultimate time-step. 
The third \define{MLP} baseline is a fully-connected multilayer perceptron (MLP) \cite{murphy2012machine} that looks at a window of earlier time-steps to predict the next time-step. 
The fourth \define{LSTM} baseline is a recurrent neural net based on the long short-term memory (LSTM) architecture \cite{hochreiter1997long}.

We also considered using a hidden Markov model (HMM) as a baseline. 
However, as Ghahramani emphasizes (\cite{ghahramani2001introduction}, Section 5), a HMM represents each of the exponential number of propositional states separately, and thus fails to generalize in the way that a first-order rule induction system does. Thus, although we did not test it, we are confident that a HMM would not perform well on our tasks.

Although the neural network architectures are very different from our system, we tried to give the various systems access to the same amount of information. 
 This means in particular that:
 \begin{itemize}
 \item
 Since our system interprets the sequence without any knowledge of the other sequences, \emph{we do not allow the neural net baselines to train on any sequences other than the one they are currently given}. Each neural net baseline is only allowed to look at the single sensory sequence it is given.
This extreme paucity of training data is unusual for data-hungry methods like neural nets, and explains their weak results. But we stress that this is the only fair comparison, given that the \sys{}, also, only has access to a single sequence.
\item
Since our system interprets the sequence without knowing anything about the relative spatial position of the sensors (it does not know, in the ECA examples, the spatial locations of the cells), we do not give the neural nets a (1-dimensional) convolutional structure, even though this would help significantly in the ECA tasks. 
\end{itemize}
The neural baselines are designed to exploit potential statistical patterns that are indicative of hidden sensor states. In the MLP baseline, we formulate the problem as a multi-class classification problem, where the input consists in a feature representation ${\bf x}$ of relevant context sensors, and a feed-forward network is trained to predict the correct state ${\bf y}$ of a given sensor in question. In the prediction task, the feature representation comprises one-hot\footnote{A one-hot representation of feature $i$ of $n$ possible features is a vector of length $n$ in which all the elements are 0 except the $i$'th element.}  representations for the state of every sensor in the previous two time steps before the hidden sensor. The training data consists of the collection of all observed states in an episode (as potential hidden sensors), together with the respective history before. Samples with incomplete history window (at the beginning of the episode) are discarded.

The MLP classifier is a 2-layer feed-forward neural network, which is trained on all training examples derived from the current episode (thus no cross-episode transfer is possible). We restrict the number of hidden neurons to (20, 20) for the two layers, respectively, in order to prevent overfitting given the limited number of training points within an episode. We use a learning rate of $10^{-3}$ and train the model using the \emph{Adam} optimiser \cite{kingma2014adam} for up to $200$ epochs, holding aside 10\% of data for early stopping. 

Given that the input is a temporal sequence, a recurrent neural network (that was designed to model temporal dynamics) is a natural choice of baseline.
But we found that the LSTM performs only slightly better than the MLP on Seek Whence tasks, and worse on the other tasks. 
The reason for this is that the extremely small amount of data (a single temporal sequence consisting of a small number of time-steps) does not provide enough information for the high capacity LSTM to learn desirable gating behaviour. The simpler and more constrained MLP with fewer weights is able to do slightly better on some of the tasks, yet both neural baselines achieve low accuracy in absolute terms.

Figure \ref{fig:baselines-chart} shows the results. Clearly, the tasks are very challenging for all four baseline systems.

\begin{figure}
\centering
\begin{tikzpicture}
\begin{axis}[
    ybar,
    enlargelimits=0.25,
    legend style={at={(0.5,-0.15)},
      anchor=north,legend columns=-1},
    ylabel={predictive accuracy},
    symbolic x coords={eca,music,Seek-Whence},
    xtick=data,
    ]
\addplot[ybar, pattern=dots] coordinates {(eca,97) (music,73) (Seek-Whence,76)};
\addplot[ybar, pattern=north west lines] coordinates {(eca,8) (music,2) (Seek-Whence,26)};
\addplot[ybar, pattern=north east lines] coordinates {(eca,29) (music,0) (Seek-Whence,33)};
\addplot[ybar, pattern=grid] coordinates {(eca,14) (music,7) (Seek-Whence,17)};
\addplot[ybar, pattern=horizontal lines] coordinates {(eca,3) (music,0) (Seek-Whence,18)};
\legend{our system (AE),constant baseline, inertia baseline, MLP baseline, LSTM baseline}
\end{axis}
\end{tikzpicture}
\caption[Comparison with baselines]{Comparison with baselines. We display predictive accuracy on the held-out final time-step.}
\label{fig:baselines-chart}
\end{figure}

Table \ref{table:baselines} shows a comparison with four baselines: a constant baseline (that always predicts the same thing), the inertia baseline (that predicts the final time-step equals the penultimate time-step), a simple neural baseline (a fully connected MLP), and a recurrent neural net (an LSTM \cite{hochreiter1997long}). The results for the neural MLP and LSTM are averaged over 5 reruns.

\begin{table}
\centering
\begin{tabular}{|l|r|r|r|}
\hline
& {\bf ECA} & {\bf Rhythm \& Music}  & {\bf Seek Whence} \\
\hline
Our system (AE) & 97.3\% & 73.3\% & 76.7\% \\
\hline
Constant baseline & 8.6\% & 2.5\%  & 26.7\% \\
\hline
Inertia baseline & 29.2\% & 0.0\% & 33.3\% \\
\hline
Neural MLP & 15.5\% & 1.3\% & 17.9\% \\
\hline
Neural LSTM & 3.3\% & 0.0\% & 18.7\% \\
\hline
\end{tabular}
\caption[Comparing our system against baselines]{Comparison with baselines. We display predictive accuracy on the held-out final time-step.}
\label{table:baselines}
\end{table}

Table \ref{table:mc-nemar-baselines} shows the McNemar test \cite{mcnemar1947note} for the four baselines. 
For each baseline, we assess the null hypothesis that its distribution is the same as the distribution of the \sys{}.
If $b$ is the proportion of tasks on which the \sys{} is inaccurate, and $c$ is the proportion of tasks in which the baseline is inaccurate, then the McNemar test statistic is 
\[
\chi^2 = \frac{(b-c)^2}{b+c}
\]

\begin{table}
\centering
\begin{tabular}{|l|r|r|r|}
\hline
& {\bf ECA} & {\bf Rhythm \& Music}  & {\bf Seek Whence} \\
\hline
AE vs constant baseline & 216.6 & 11.9  & 7.8 \\
\hline
AE vs inertia baseline & 164.2 & 12.7 & 6.3 \\
\hline
AE vs neural MLP & 200.6 & 11.1 & 7.0 \\
\hline
AE vs neural LSTM & 242.1 & 12.7 & 6.9 \\
\hline
\end{tabular}
\caption[The McNemar test comparing our system to each baseline]{The McNemar test comparing our system (AE) to each baseline. The McNemar test statistic generates a $\chi^2$ distribution with 1 degree of freedom. For each entry in the table, the null hypothesis (that the baseline's distribution is the same as our system's distribution) is extremely unlikely.}
\label{table:mc-nemar-baselines}
\end{table}

\subsection{Our system handles retrodiction and imputation just as easily as prediction}
\label{sec:retrodiction-imputation}

To verify that our system is just as capable of retrodicting earlier values and imputing missing intermediate values as it is at predicting future values, we ran tests where the unseen hidden sensor values were at the first time step (in the case of retrodiction) or randomly scattered through the time-series (in the case of imputation).
We made sure that the number of hidden sensor values was the same for prediction, retrodiction, and imputation.

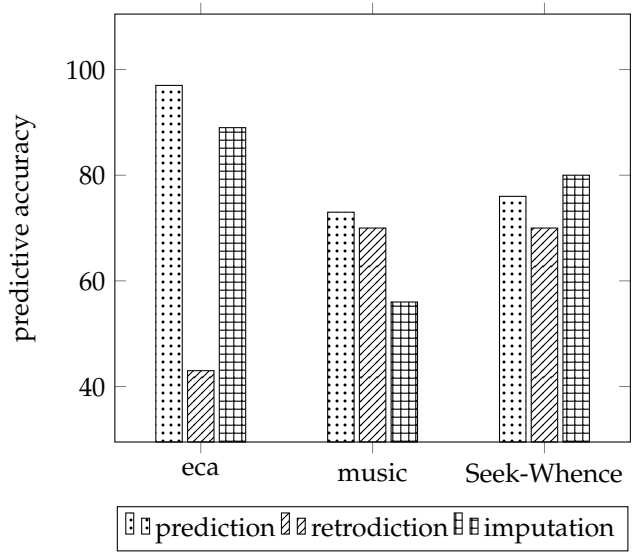
\begin{figure}
\centering
\begin{tikzpicture}
\begin{axis}[
    ybar,
    enlargelimits=0.25,
    legend style={at={(0.5,-0.15)},
      anchor=north,legend columns=-1},
    ylabel={predictive accuracy},
    symbolic x coords={eca,music,Seek-Whence},
    xtick=data,
    ]
\addplot[ybar, pattern=dots] coordinates {(eca,97) (music,73) (Seek-Whence,76)};
\addplot[ybar, pattern=north east lines] coordinates {(eca,43) (music,70) (Seek-Whence,70)};
\addplot[ybar, pattern=grid] coordinates {(eca,89) (music,56) (Seek-Whence,80)};

\legend{prediction,retrodiction,imputation}
\end{axis}
\end{tikzpicture}
\caption[Comparing prediction with retrodiction and imputation]{Comparing prediction with retrodiction and imputation. In retrodiction, we display accuracy on the held-out initial time-step. In imputation, a random subset of atoms are held-out; the held-out atoms are scattered throughout the time-series. In other words, there may be different held-out atoms at different times. The number of held-out atoms in imputation matches the number of held-out atoms in prediction and retrodiction.}
\label{fig:retrodiction-and-imputation-chart}
\end{figure}

\begin{figure}
\begin{center}
\begin{tikzpicture}[b/.style={draw, minimum size=3mm,   
       fill=black},w/.style={draw, minimum size=3mm},
       m/.style={matrix of nodes, column sep=1pt, row sep=1pt, draw, label=below:#1}, node distance=1pt]

\matrix (A) [m=0]{
|[w]|&|[w]|&|[w]|&|[w]|&|[w]|&|[b]|&|[w]|&|[w]|&|[w]|&|[w]|&|[w]|\\
|[w]|&|[w]|&|[w]|&|[w]|&|[w]|&|[w]|&|[w]|&|[w]|&|[w]|&|[w]|&|[w]|\\
|[w]|&|[w]|&|[w]|&|[w]|&|[w]|&|[w]|&|[w]|&|[w]|&|[w]|&|[w]|&|[w]|\\
|[w]|&|[w]|&|[w]|&|[w]|&|[w]|&|[w]|&|[w]|&|[w]|&|[w]|&|[w]|&|[w]|\\
};
\end{tikzpicture}  
\end{center}
\caption[One trajectory for ECA rule \# 0]{One trajectory for ECA rule \# 0. This trajectory shows how information is lost as we progress through time. Here, clearly, retrodiction (where the first row is held-out) is much harder than prediction (where the final row is held-out).}
\label{fig:trajectory0}
\end{figure}
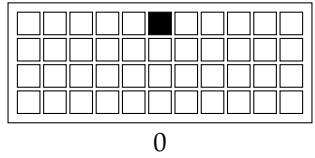

Figure \ref{fig:retrodiction-and-imputation-chart} shows the results.
The results are significantly lower for retrodiction in the ECA tasks, but otherwise comparable.
The reason for retrodiction's lower performance on ECA is that for a particular initial configuration there are a significant number (more than 50\%) of the ECA rules that wipe out all the information in the current state after the first state transition, and all subsequent states then remain the same. So, for example, in Rule \# 0, one trajectory is shown in Figure \ref{fig:trajectory0}. Here, although it is possible to predict the future state from earlier states, it is not possible to retrodict the initial state given subsequent states.


The results for imputation are comparable with the results for prediction. Although the results for rhythm and music are lower, the results on Seek Whence are slightly higher (see Figure \ref{fig:retrodiction-and-imputation-chart}).

\subsection{The features of our system are essential to its performance}
\label{sec:ablation}

To verify that the unity conditions are doing useful work, we performed a number of experiments in which the various conditions were removed, and compared the results.
We ran four ablation experiments.
In the first, we removed the check that the theory's trace covers the input sequence.
In the second, we removed the check on conceptual unity.
Removing this condition means that the unary predicates are no longer connected together via exclusion relations $\oplus$, and the binary predicates are no longer constrained by $\exists !$ conditions.
In the third ablation test, we removed the check on spatial unity.
Removing this condition means allowing objects which are not connected via binary relations.
In the fourth ablation test, we removed the cost minimization part of the system.
Removing this minimization means that the system will return the first interpretation it finds, irrespective of size.

The results of the ablation experiments are displayed in Table \ref{table:ablation}.

The first ablation test, where we remove the check that the generated sequence of sets of ground atoms respects the original sensory sequence, performs very poorly. 
Of course, if the generated sequence does not cover the given part of the sensory sequence, it is highly unlikely to accurately predict the held-out part of the sensory sequence. 
This test is just a sanity check that our evaluation scripts are working as intended.

The second ablation test, where we remove the check on conceptual unity, also performs poorly.
The reason is that without constraints, there are no incompossible atoms.
Recall that two atoms are incompossible if there is some $\oplus$ constraint or some $\exists !$ constraint that means the two atoms cannot be simultaneously true.
But  the frame axiom forces an atom that was true at the previous time-step to also be true at the next time-step unless the old atom is incompossible with some new atom:
we add $\alpha$ to $H_t$ if $\alpha$ is in $H_{t-1}$ and there is no atom in $H_t$ that is incompossible with $\alpha$.
But if there are no incompossible atoms, then all previous atoms are always added.
Therefore, if there are no $\oplus$ and $\exists !$ constraints, then the set of true atoms monotonically increases over time.
This in turn means that state information becomes meaningless, as once something becomes true, it remains always true, and cannot be used to convey information. 

\begin{table}
\centering
\begin{tabular}{|l|r|r|r|}
\hline
& {\bf ECA} & {\bf Rhythm \& Music}  & {\bf Seek Whence} \\
\hline
Full system (AE) & 97.3\% & 73.3\% & 76.7\% \\
\hline
No check that $S \sqsubseteq \tau(\theta)$ & 5.1\% & 3.0\% & 4.6\% \\
\hline
No conceptual unity & 5.3\% & 0.0\% & 6.7\% \\
\hline
No spatial unity & 95.7\% & 73.3\% & 73.3\% \\
\hline
No cost minimization & 96.7\% & 56.6\% & 73.3\% \\
\hline
\end{tabular}
\caption[Ablation experiments]{Ablation experiments. We display predictive accuracy on the final held-out time-step.}
\label{table:ablation}
\end{table}


When we remove the spatial unity constraint, the results for the rhythm tasks are identical, but the results for the ECA and Seek Whence tasks are lower. 
The reason why the results are identical for the rhythm tasks is because the background knowledge provided (the $r$ relation on notes, see Section \ref{sec:rhythms-and-tunes}) means that the spatial unity constraint is guaranteed to be satisfied. 
The reason why the results are lower for ECA tasks is because interpretations that fail to satisfy spatial unity contain disconnected clusters of cells (e.g.~cells $\{c_1, ..., c_5\}$ are connected by $r$ in one cluster, while cells $\{c_6, ..., c_{11}\}$ are connected in another cluster, but  $\{c_1, ..., c_5\}$ and $\{c_6, ..., c_{11}\}$ are disconnected). Interpretations with disconnected clusters tend to generalize poorly and hence predict with less accuracy. 
The reason why the results are only slightly lower for the Seek Whence tasks is because the lowest cost unified interpretation for most of these tasks also happens to satisfy spatial unity. 
We have assumed, in deference to Kant, that the spatial unity constraint does useful work.
But it is at least conceptually possible that this constraint is not needed in at least some domains.
In future work, we shall test the \sys{} in a much wider variety of domains, to understand when spatial unity is\footnote{Recently, we have found other cases where this spatial unity constraint is necessary. Andrew Cropper has some recent unpublished work using object invention in which the spatial unity constraint was found to be essential.} and is not\footnote{Some philosophers (e.g.~Strawson \cite{strawson2018bounds}) have questioned whether the spatial unity constraint is, in fact, necessary.} important.

The results for the fourth ablation test, where we remove the cost minimization, are broadly comparable with the full system in ECA and Seek Whence, but are markedly worse in the rhythm / music tasks. 
But even if the results were comparable in all tasks, there are independent reasons to want to minimize the size of the interpretation, since shorter interpretations are more human-readable.
On the other hand, it is significantly more expensive to compute the lowest cost theory than it is to just find any unified theory. So in some domains, where the difference in accuracy is minimal, the cost minimization step can be avoided.

\subsection{A comparison with ILASP}
\label{sec:ilasp}

In order to assess the efficiency of our system, we compared it to ILASP \cite{law2014inductive,law2015learning,law2016iterative,law2018complexity}, a state of the art Inductive Logic Programming algorithm\footnote{Strictly speaking, ILASP is a family of algorithms, rather than a single algorithm. We used ILASP2 \cite{law2015learning} in this evaluation. We are very grateful to Mark Law for all his help in this comparative evaluation.}.

Unlike traditional ILP systems that learn definite logic programs, ILASP learns \emph{answer set programs}\footnote{Answer set programming under the stable model semantics is distinguished from traditional logic programming in that it is purely declarative and each program has multiple solutions (known as answer sets). Because of its non-monotonicity, ASP is well suited for knowledge representation and common-sense reasoning \cite{mueller2014commonsense,gelfond2014knowledge}.}.
ILASP is a powerful and general framework for learning answer set programs; it is able to learn choice rules, constraints, and even preferences over answer sets \cite{law2015learning}.
%
%

Because of the generality of the Learning from Answer Sets framework, we can express an apperception task within it.
Of course, since ILASP was not designed specifically with this task in mind, there is no reason it would be as efficient as a program synthesis technique which was targeted specifically at apperception tasks.

In ILASP, the set of $H$ of hypothesis clauses is defined by a set of \define{mode declarations}: statements specifying the sort of atoms that are allowed in the heads and bodies of clauses.
For example, the declaration \verb|#modeh(p(var(t1)))| states that an atom $p(X)$ can appear in the head of a clause, where $X$ is some variable of type \verb|t1|.
The declaration \verb|#modeb(2, r(var(t1), const(t2)))| states that an atom $f(X, k)$ can appear in the body of a clause, where $k$ is a constant of type \verb|t2|. The parameter 2 in the \verb|modeb| declaration specifies that an atom of this form can appear at most two times in the body of any rule.

ILASP uses a similar approach to ASPAL \cite{corapi,corapi2} to induce programs: it uses the mode declarations to generate a large set of hypothesis clauses, and adds a clause flag to each clause indicating whether or not the clause is operative. 
Then, the induction problem is transformed into the simpler problem of deciding which clause flags to turn on.
%
To generate an interpretation for an apperception task, we need to generate a set of initial atoms, a set of static rules and a set of causal rules. 
We generate mode declarations for each type.
Each potential initial atom \verb|X| is turned into a \verb|modeh| declaration \verb|#modeh(init(X))|.
Static rules and causal rules are generated by \verb|#modeb| and \verb|#modeh| declarations specifying the atoms allowed in the body and head of each rule.

\subsubsection{Evaluation}

ILASP is able to solve some simple apperception tasks. For example, ILASP is able to solve the task in Example \ref{example:one}. 
But for the ECA tasks, the music and rhythm tasks, and the Seek Whence tasks, the ASP programs generated by ILASP were not solvable because they required too much memory.

In order to understand the memory requirements of ILASP on these tasks, and to compare our system with ILASP in a fair like-for-like manner, we looked at the size of the grounded ASP programs.
Recall that both our system and ILASP generate ASP programs that are then grounded (by \verb|gringo|) into propositional clauses that are then solved (by \verb|clasp|).\footnote{These two programs are part of the Potassco ASP toolset: https://potassco.org/clingo/}
The grounding size determines the memory usage and is strongly correlated with solution time.

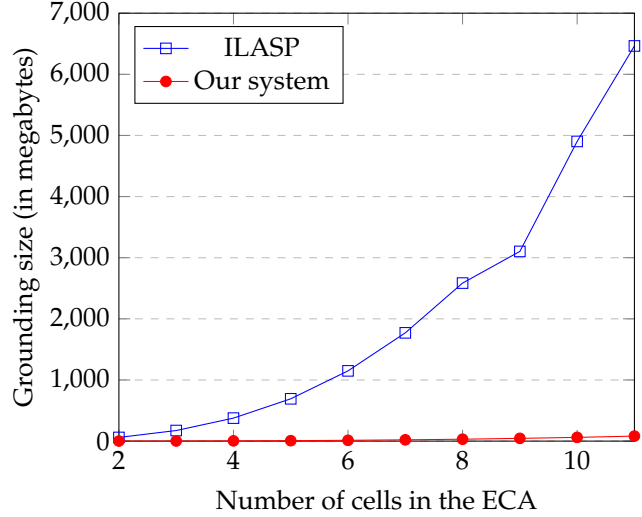
\begin{figure}
\begin{center}
\begin{tikzpicture}
\begin{axis}[
    title={},
    xlabel={Number of cells in the ECA},
    ylabel={Grounding size (in megabytes)},
    xmin=2, xmax=11,
    ymin=0, ymax=7000,
    xtick={2,4,6,8,10},
    ytick={0,1000,2000,3000,4000,5000,6000,7000},
    legend pos=north west,
    ymajorgrids=true,
    grid style=dashed,
]
 
\addplot[
    color=blue,
    mark=square,
    ]
    coordinates {
    (2,60)(3,173)(4,376)(5,692)(6,1149)(7,1771)(8,2585)(9,3103)(10,4902)(11,6464)
    };
    \addlegendentry{ILASP}

\addplot[
    color=red,
    mark=*,
    ]
    coordinates {
    (2,1)(3,2)(4,4)(5,7)(6,13)(7,21)(8,31)(9,45)(10,61)(11,82)
    };
    \addlegendentry{Our system}
 
\end{axis}
\end{tikzpicture}
\end{center}
\caption{Comparing our system and ILASP w.r.t. grounding size}
\label{fig:grounding-comparison}
\end{figure}

We took a sample ECA, Rule 245, and looked at the grounding size as the number of cells increased from 2 to 11.
The results are in Table \ref{table:grounding-comparison} and Figure \ref{fig:grounding-comparison}.

\begin{table}
\centering
\begin{tabular}{|l|r|r|}
\hline
{\bf \# cells} & {\bf Our System} & {\bf ILASP} \\
\hline
2 & 0.6 & 60.7 \\
\hline
3 & 1.8 & 173.7 \\
\hline
4 & 4.0 & 376.0 \\
\hline
5 & 7.8 & 692.8 \\
\hline
6 & 13.4 & 1149.6 \\
\hline
7 & 21.3 & 1771.9 \\
\hline
8 & 31.7 & 2585.1 \\
\hline
9 & 45.1 & 3103.4 \\
\hline
10 & 61.8 & 4902.6 \\
\hline
11 & 82.6 & 6464.1 \\
\hline
\end{tabular}
\caption[Like-for-like comparison between our system and ILASP]{Like-for-like comparison between our system and ILASP. We compare the size of the ground programs (in megabytes) generated by both systems as the number of cells in the ECA increases from 2 to 11.}
\label{table:grounding-comparison}
\end{table}

As we increase the number of cells, the grounding size of the ILASP program grows much faster than the corresponding  \sys{} program. 
The reason for this marked difference is the different ways the two approaches represent rules.
In our system, rules are interpreted by an interpreter that operates on reified representations of rules.
In ILASP, by contrast, rules are \emph{compiled} into ASP rules.
This means, if there are $|U_\phi|$ unground atoms and there are at most $N_B$ atoms in the body of a rule, then ILASP will generate $|U_\phi|^{N_B+1}$ different clauses.
When it comes to grounding, if there are $|\Sigma_\phi|$ substitutions and $t$ time-steps, then ILASP will generate at most
$|U_\phi|^{N_B+1} \cdot |\Sigma_\phi| \cdot t$ ground instances of the generated clauses.
Each ground instance will contain $N_B + 1$ atoms, so there are $(N_B + 1) \cdot |U_\phi|^{N_B+1} \cdot |\Sigma_\phi| \cdot t$ ground atoms in total.

Compare this with our system. 
Here, we do not represent every possible rule explicitly as a separate clause.
Rather, we represent the possible atoms in the body of a rule by an ASP choice rule.
If there are $N_\rightarrow$ static rules and $N_{\fork}$ causal rules, then this choice rule only generates $N_\rightarrow + N_{\fork}$ ground clauses, each containing $|U_\phi|$ atoms. 

The reason why our system has such lower grounding sizes than ILASP is because  ILASP considers \emph{every possible} subset of the hypothesis space, while our system (by restricting to at most $N_{\rightarrow} + N_{\fork}$ rules) only considers \emph{subsets of length at most $N_{\rightarrow} + N_{\fork}$}.

\section{Conclusion}
\label{sec:conclusion}

This paper describes the various experiments we have performed to evaluate the \sys{} in a variety of domains.
in each domain, we tested its ability to predict future values, retrodict previous values, and impute missing intermediate values. 
Our system achieves good results across the board, outperforming neural network baselines and also state of the art ILP systems.

Of particular note is that the \sys{} is able to achieve human performance on challenging sequence induction intelligence tests.
We stress, once more, that the system was not hard-coded to solve these tasks.
Rather, it is a general \emph{domain-independent} sense-making system that is able to apply its general architecture to the particular problem of Seek Whence induction tasks, and is able to solve these problems ``out of the box'' without human hand-engineered help.
We also stress, again, that the system did not learn to solve these sequence induction tasks by being presented with hundreds of training examples\footnote{Barrett et al \cite{barrett2018measuring} train a neural network to learn to solve Raven's progressive matrices from thousands of training examples.}. 
Indeed, the system had never seen a \emph{single} such task before.
Instead, it applied its general sense-making urge to each individual task, \emph{de novo}.
We also stress that the interpretations produced are human readable and can be used to provide explanations and justifications of the decisions taken:
when the \sys{} produces an interpretation, we can not only see what it predicts will happen next, but we can also understand \emph{why} it thinks this is the right continuation. 
We believe these results are highly suggestive, and shows that a sense-making component such as this will be a key aspect of any general intelligence.

Our architecture, an unsupervised program synthesis system, is a purely symbolic system, and 
as such, it inherits two key advantages of ILP systems \cite{evans2018learning}.
First, the interpretations produced are \emph{interpretable}. 
Because the output is symbolic, it can be read and verified by a human\footnote{Large machine-generated programs are not always easy to understand. But machine-generated symbolic programs are certainly easier to understand than the weights of a neural network. See Muggleton et al \cite{muggleton2018ultra} for an extensive discussion. }.
Second, it is very \emph{data-efficient}.
Because of the language bias of the \logic{} language, and the strong inductive bias provided by the unity conditions, the system is able to make sense of extremely short sequences of sensory data, without having seen any others.

However, the system in its current form has some limitations that we wish to make explicit.
First, the sensory input must be discretized before it can be passed to the system.
We assume some prior system has already discretized the continuous sensory values by grouping them into buckets.
Second, our implementation as described above assumes all causal rules are fully deterministic.
Third, the size of the search space means that our system is currently restricted to small-to-medium-size problems.\footnote{This is not because our system is carelessly implemented: the \sys{} is able to synthesize significantly larger programs than state-of-the-art ILP systems (see the comparison with ILASP in Section \ref{sec:ilasp}).}
We are evaluating a number of alternative approaches to program synthesis in order to scale up to larger problems.\footnote{In particular, we believe that some of the techniques used in Popper \cite{cropper2020learning} will allow us to scale to significantly larger and harder problems.}
Going forward, we believe that the right way to build complex theories is incrementally, using curriculum learning: the system should consolidate what it learns in one episode, storing it as background knowledge, and reusing it in subsequent episodes.

We hope in future work to address these limitations. 
But we believe that, even in its current form, the \sys{} shows considerable promise as a prototype of what a general-purpose domain-independent sense-making machine must look like.


\section*{References}

\bibliography{main}

\end{document}